\documentclass[10pt]{article}

\usepackage{graphicx}%
\usepackage{multirow}%
\usepackage{amsmath,amssymb,amsfonts}%
\usepackage{amsthm}%
\usepackage{mathrsfs}%
\usepackage[title]{appendix}%
\usepackage{xcolor}%
\usepackage{textcomp}%
\usepackage{manyfoot}%
\usepackage{booktabs}%
\usepackage{algorithm}%
\usepackage{algorithmicx}%
\usepackage{algpseudocode}%
\usepackage{listings}%
\usepackage{rotating}
\usepackage{nicefrac}
\usepackage{bm}
\usepackage{bbm}
\usepackage{subcaption}

\DeclareGraphicsExtensions{.pdf}

\usepackage{hyperref}

\catcode`\.=\active\def.{\char'56\allowbreak}\catcode`\.=12
\catcode`\/=\active\def/{\char'57\allowbreak}\catcode`\/=12
\catcode`\:=\active\def:{\char'72\allowbreak}\catcode`\:=12  
\catcode`\@=\active\def@{\char'100\allowbreak}\catcode`\@=12 
\catcode`\-=\active\def-{\char'55\allowbreak}\catcode`\-=12  
\def\URL{\bgroup\catcode`\.=\active\catcode`\/=\active\catcode`\:=\active\catcode`\@=\active\catcode`\-=\active\def~{\char126}\tt\URLaux}
\def\URLaux#1{#1\egroup}

\date{}

\begin{document}

\title{A Performance Analysis of Basin Hopping Compared to Established Metaheuristics for Global Optimization}
\author{%
Marco Baioletti\thanks{Department of Mathematics and Computer Science, University of Perugia, Italy} \and 
Valentino Santucci\thanks{Department of Humanities and International Social Sciences, University for 
Foreigners of Perugia, Italy} \and 
Marco Tomassini\thanks{Department of Information Systems, University of Lausanne, Switzerland} 
}
\maketitle

\begin{abstract}
During the last decades many metaheuristics for global numerical optimization have been proposed. Among them, Basin Hopping is very simple and straightforward to implement, although rarely used outside its original Physical Chemistry community. In this work, our aim is to compare Basin Hopping, and 
two population variants of it,
with readily available implementations of the well known metaheuristics Differential Evolution, Particle Swarm 
Optimization, and Covariance Matrix Adaptation Evolution Strategy. We perform  numerical experiments using the \textit{IOH profiler} environment with the BBOB test function set and two difficult real-world problems. The experiments were carried out in two different but complementary ways: by measuring the performance under a fixed budget of function evaluations and by considering a fixed target value. The general conclusion is that 
Basin Hopping and 
its newly introduced population variant
are almost as good as Covariance Matrix Adaptation on the synthetic benchmark functions and better than it on the two hard cluster energy minimization problems. Thus, the proposed analyses show that Basin Hopping can be considered a good candidate for global numerical optimization problems along with the more established metaheuristics, especially if one wants to obtain quick and reliable results on an unknown problem.
\noindent
\end{abstract}


\section{Introduction}
\label{sec:intro}

The goal in global function optimization is the maximization or
minimization of an arbitrary function, possibly subject to some constraints. Global optimization problems arise
very often in many fields such as natural sciences, engineering, and machine learning among others.
The problem of global minimization involves finding the minimum value $m$ of a function $f$ within a domain~\raisebox{2pt}{$\chi$}:

\begin{equation*}
\begin{aligned}
&\underset{\textbf{x}}{\min}
& & \hspace{-0.3cm} \{f(\textbf{x}) \;:\; \textbf{x} \in \raisebox{2pt}{$\chi$}\}  \\
\end{aligned}
\end{equation*}

Typically, it is also desirable to determine the argument or the point, or set of points, $\textbf{x} \in \raisebox{2pt}{$\chi$}$ that yields the minimum value $m$ provided by the function:

\begin{equation*}
\begin{aligned}
& \underset{\textbf{x}}{\arg\min}
& &  \hspace{-0.3cm} \{\textbf{x} \in \raisebox{2pt}{$\chi$} \; : \; f(\textbf{x}) = m \},  \\
\end{aligned}
\end{equation*}

where $\textbf{x}$ is a column vector of real-valued variables \mbox{$[x_1, x_2, \ldots, x_D]^T \in \raisebox{2pt}{$\chi$}$} of dimension $D$ and $\raisebox{2pt}{$\chi$}$ is a subset of $\mathbb{R}^D$ that defines the feasible set containing any solution $\textbf{x}$. The definition can be adapted for constrained optimization problems with an appropriate definition of $\raisebox{2pt}{$\chi$}$, and maximizing the objective function is accomplished by replacing $f(\textbf{x})$ with $-f(\textbf{x})$. We will use the simplest constraints, called box constraints, which restrict variables to belong to a hyperrectangle defined by \mbox{$[a_1,b_1] \times \cdots \times [a_i,b_i] \times \cdots \times [a_D,b_D]$}, where $a_i$ and $b_i$ represent, respectively, the lower and upper bounds of variable $x_i$.

Many algorithms have been devised over several decades
to solve the global optimization problem. However, most of them are limited to
the restricted class of convex functions, for which a locally optimal
solution is also globally optimal. 
However, in numerous significant scenarios, optimization problems are not linear or convex. Rather, they are frequently non-linear, discontinuous, highly multimodal, or even non-differentiable. Moreover, it is often the case that there is no analytical expression for the function being optimized, and its value is instead obtained from a measurement or simulation. This is known as a ``black box'' scenario~\cite{audet2017}, which necessitates the use of algorithms that exclusively rely on function values.
In order to solve, at least approximately, this wider class of global optimization problems
new heuristic methods, often inspired by analogies with natural phenomena, have been devised.
Among the many examples, the most well known and successful have been
Evolution Strategies (ES)~\cite{back2013}, Differential Evolution~(DE)~\cite{storn1997},
Simulated Annealing (SA)~\cite{SA}, and Particle Swarm Optimization (PSO)~\cite{kennedy1995}. Most of these algorithms rely on populations of candidate solutions while others, 
such as Simulated Annealing, are trajectory-based. 
These approaches, collectively known as \textit{metaheuristics}, exhibit varying and, at times, uncertain convergence behavior, and do not offer any assurance of global optimality. However, they all share a crucial attribute of exploring the search space globally, potentially enabling them to avoid local optima. Additionally, metaheuristics are beneficial to practitioners since they do not necessitate a comprehensive understanding of the mathematical and problem-related aspects, making them relatively easy to comprehend and implement.

It must be said that there are also several others more mathematically rigorous algorithms for
global optimization (see, e.g.~\cite{liberti,locatelli2013global,kochenderfer2019}). However, they require more specialized
technical knowledge to be used by the non-specialist. In spite of their importance, we shall not pursue them
further here but, for the interested reader, we point out a recent comparison between metaheuristics
and mathematically rigorous methods on a large benchmark function set under a limited budget of
function evaluations~\cite{sergeyev2018}.
We also note a previous investigation by Pham and Castellani which had similar objectives as the present study~\cite{pham2014}. The authors compare ES, PSO, the Bees Algorithm, and Artificial Bee Colony among them. However, they do not include
Basin Hopping, DE, and Covariance Matrix Adaptation Evolution Strategy (CMA-ES), use a different benchmark suite, and do not consider real-world problems. 

Basin Hopping (BH) originated in computational Chemical Physics to search for minimum energy states of atomic clusters and biological macromolecules, e.g.,~\cite{wales1997,wales1999,doye2004}
and it is still actively pursued in those fields for solving difficult global 
optimization problems see, e.g.,~\cite{BHkucharik2014,zhou2019,2021crystal}. Over the years the algorithm has included
an increasing amount of chemical knowledge and constraints thus becoming more specialized for use in these particular
fields. However, in its original structure and simplicity it can be used for general global function optimization
and it is from this point of view that we shall consider it here.
Indeed, as far as we know Basin Hopping has never been 
systematically compared to other metaheuristics before and our aim is to find out how BH fares with respect
to three well known and commonly used metaheuristics: Differential
Evolution, Covariance Matrix Adaptation, and Particle Swarm Optimization. 
In addition, we also propose and test a new population-based basin hopping implementation that we call BHPOP.
A preliminary version of the study has appeared in~\cite{baioletti2022}  but here we extend and complete it in
several ways as follows.
We introduce two new metaheuristics (CMA-ES and BHPOP) in the comparison, one of which is new (BHPOP); we
provide a fixed-target analysis besides the fixed-budget
analysis together with a new and more complete statistical analysis of the results. Finally, we add a performance 
analysis of two hard real-world problems. 
For the comparison the recent and widely used BBOB benchmark test suite was employed~\cite{hansen2009real}. The BBOB test functions are provided in the 
\textit{IOH Profiler} environment~\cite{doerr2018iohprofiler} which is well organized and
easy to use.
The only previous study we know of testing BH on a modern benchmark suite 
is~\cite{baudisBH}. In this paper the author tests BH from the Python SciPy library in an algorithm portfolio context 
using the same BBOB benchmark suite that we use here but, instead of comparing different metaheuristics as
we do, the author focuses on testing several local optimizers within BH itself using only three relatively small problem dimensions.  Our study is much more extensive, has different objectives and,
moreover, in order to test the set of metaheuristics under more practical conditions, 
we also included in the comparison several instances of a couple of difficult real world optimization problems. It is important to point out
that our aim is not to reach the best possible performances. Rather, we use relatively standard off-the-shelf versions of the
algorithms instead of the best versions available and we address robustness issues, in the sense of preferring
uniformly good performances and ease of use. Our results should thus be useful to practitioners wanting to tackle a new problem
easily and effectively, before spending more resources with specialized techniques.

The structure of the article is the following. The next section provides a brief introduction to the various metaheuristics
to be compared, including BHPOP, our population version of BH.
Sect.~\ref{sec:expsetup} describes the experimental setup, including the benchmarking environment, the set of test functions,
and the results obtained. 
Sect.~\ref{sec:expres} gives a general discussion of the results and of their significance, while Sect.~\ref{sec:realworld} presents experimental results on real-world problems. Finally, Sect.~\ref{sec:concl} presents our conclusions.

\section{Metaheuristics}
\label{sec:mh}

For the sake of self-containedness, in this section we briefly describe the compared metaheuristics, with an emphasis on Basin Hopping, which is customarily used in Chemical Physics blue but has also been effective on other types of global optimization 
problems~\cite{grosso2007-ex,grosso2007,locatelli2013global}.
We also introduce BHPOP, i.e., our own population-based variant of BH.

\subsection{Basin Hopping} 
\label{bhop}

A synopsis of the BH algorithm (see~\cite{locatelli2013global} for more details) is given in pseudocode~\ref{bh}, in which solutions
$\textbf{s},\textbf{x},\textbf{y},\textbf{z}$ are to be considered as $D$-dimensional vectors.

\begin{algorithm}
\caption{Basin Hopping}
\label{bh}
\begin{algorithmic}[0]
\State $\textbf{s} \leftarrow$ initial solution
\State $\textbf{x} \leftarrow $ minimize($f,\textbf{s}$)
\While{termination condition is false}
\State $\textbf{y} \leftarrow $ perturbation($\textbf{x}$)
\State $\textbf{z} \leftarrow $ minimize($f,\textbf{y}$)
\State $\textbf{x} \leftarrow $ acceptance test($\textbf{x}, \textbf{z}$)
\EndWhile
\State \textbf{return} $\textbf{x}, f(\textbf{x})$
\end{algorithmic}
\end{algorithm}

An initial solution $\textbf{s}$ is created randomly or heuristically. This solution must lie within the basin of attraction of some local optimum, whose coordinates $\textbf{x}$ can be discovered by conducting a local search procedure (minimize) starting from $\textbf{s}$. The algorithm then proceeds through three stages in each iteration. Firstly, the current solution $\textbf{x}$ undergoes some form of coordinate change, resulting in a new solution $\textbf{y}$. Following that, the local minimizer finds the new local minimum $\textbf{z}$, starting at $\textbf{y}$. There are two possible outcomes: either $\textbf{z}$ differs from $\textbf{x}$, indicating that the algorithm has successfully escaped from the basin of attraction of $\textbf{x}$, or it is identical to $\textbf{x}$. In the latter case, the perturbation was not enough, and $\textbf{y}$ belongs to the original basin of attraction, causing the search to rediscover the same minimum. Finally, the acceptance stage entails determining whether the new solution $\textbf{z}$ is accepted as the starting point for the subsequent cycle. If $f(\textbf{z})=f(\textbf{x})$, the search resumes by attempting another perturbation from the starting point $\textbf{x}$. Otherwise, $\textbf{z}$ is accepted unconditionally or subject to some condition.

Basin Hopping accepts the new solution based on a Monte Carlo test, which is similar to simulated annealing except that the temperature is kept constant~\cite{wales1999}. If the new solution $\textbf{z}$ is better than $\textbf{x}$, it is always accepted. However, if it is worse than $\textbf{x}$, the acceptance is based on the probability $\exp\left({-\beta(f(\textbf{z}) - f(\textbf{x}))}\right)$, where $\beta$ is a parameter inversely proportional to a simulated temperature.
In the present work, we use the \textit{Monotonic Sequence Basin Hopping} in which
a solution $\textbf{z}$ is accepted if and only if $f(\textbf{z}) < f(\textbf{x})$~\cite{leary2000}. This further simplifies the algorithm
leaving a single free parameter to be set, the perturbation strength.
As is commonly done in metaheuristic approaches to global optimization, the algorithm terminates
after a fixed number of iterations or after a predetermined computing time, or when there is no
improvement in objective function value within a certain precision during a given number of iterations.
It is worth to notice that the best solution found by BH is always a local minimum. In the monotonic sequence version of BH it also corresponds to the current solution \textbf{x}. Many other optimization algorithms are not able to guarantee this important property.

Therefore, the effectiveness of the Basin Hopping algorithm relies on the proper coordination of its three main elements: the local minimization procedure, the perturbation technique, and the acceptance criterion. Achieving a good synergy among these components is crucial for the search efficiency. However, the perturbation technique can be challenging to design. If the perturbations are too small compared to the problem's typical basin size, the search may frequently return to the initial basin, leading to reduced efficiency. Conversely, if the jumps are too extensive, the search may turn into a random walk in the solution space, which is also an inefficient strategy. Additionally, each function has its unique landscape, which is generally unknown unless the function space is sampled beforehand or during the search.
In this work the perturbation corresponds to the addition of a random vector of small magnitude to the current solution $\textbf{x}$. 
Formally, given the domain intervals $[a_j, b_j]$, for $j=1,\ldots,D$, of the problem at hand, together with the scale factor $s=\frac{1}{10}$, we create a perturbation vector $\sigma \in \mathbb{R}^D$, such that each coordinate $\sigma_j$ is independently sampled uniformly at random in the zero-centered interval $\left[- s \frac{(b_j-a_j)}{2}, s \frac{(b_j-a_j)}{2}\right]$, then the perturbed solution is calculated as $\textbf{x} + \sigma$.
We have also seen that there are various ways of implementing the acceptance phase that impose different intensification/diversification
ratios, thus influencing speed search and its convergence. 

In spite of its apparent simplicity, the real core of BH is the local minimization phase which is technically
difficult if one wants to avoid numerical and other errors.  Fortunately,
reliable mathematical minimization algorithms  have been developed over several decades and are now 
routinely available in software libraries. Quasi-Newton methods are particularly efficient and we
use a local minimization method called ``L-BFGS-B'' in SciPy, an improvement of the Broyden-Fletcher-Goldfarb-Shanno (BFGS) algorithm~\cite{shanno}. 
It is worth remarking at this point that BH also works for
non-continuous or non-differentiable functions in a black-box environment. For instance, if analytical derivatives are not provided, or are unknown, BFGS can approximate them with finite differences. Moreover, local search can be performed with any working minimization algorithm, for example, the Nelder-Mead algorithm~\cite{nelder1965}, 
Powell's method~\cite{powell1964}, or any other derivative-free local descent method such as one of those described in~\cite{audet2017}.

Finally, it is perhaps useful to point out that BH can be seen as a continuous analogue of the
algorithm called Iterated Local Search (ILS) (see, for instance,~\cite{lourenco2019} and references therein) 
which is often used for difficult combinatorial optimization problems. Both algorithms do a walk over the local minima of the search space but they use fundamentally different perturbations and local optimizers. In spite of their similarity, 
ILS and BH seem to have been developed at about the same time but independently of each other.

\subsection{Population-based Basin Hopping} 
\label{pbh}

The original BH algorithm is a trajectory-based metaheuristic but there is nothing in it that prevents multiple
searchers to be used. We have thus implemented a population-based BH, to which we refer with the name BHPOP. 

Unknown to us, we recently realized that there existed a previous version of Basin Hopping using a population proposed by Grosso et al.~\cite{grosso2007,grosso2007-ex}, which they called PBH. Our independent implementation is different from the one described in~\cite{grosso2007,grosso2007-ex} in its selection and reproduction phase. 
It is interesting to compare the two algorithms and, before doing so in the experimental part, we describe both algorithms at the high level by giving their respective pseudocodes.

Algorithm~\ref{alg:pbh} describes our version, namely BHPOP.
At the beginning we randomly
generate a population of $N$ points in the function domain and locally minimize their function values. Then the
algorithms enters a loop where, at each iteration, a solution from the current population is randomly chosen with a probability proportional to its fitness with the standard
roulette wheel method and subjected to perturbation and to a local minimization phase. 
If the new minimum $\textbf{z}$ is better than the worst element in the current population, $\textbf{z}$ enters the population replacing it. 
The loop is executed until a termination condition is met. 

\begin{algorithm}
\caption{Population-Based Basin Hopping}
\label{alg:pbh}
\begin{algorithmic}[0]
\State randomly initialize the $N$ solutions $\{\textbf{x}\}_{1}^{N}$ in the population $P$
\State locally minimize all the solutions $\{\textbf{x}\}_{1}^{N}$ in $P$
\State set $\textbf{x}_\text{worst}$ to the point with the maximum function value found
\While{termination condition is false}
	\State choose $\textbf{x}$ in $P$ with the fitness-proportionate method
	\State $\textbf{y} \leftarrow$  perturb($\textbf{x}$)
	\State $\textbf{z} \leftarrow$  minimize(f($\textbf{y})$)
	\If{$f(\textbf{z}) < f(\textbf{x}_\text{worst})$}
	\State delete $\textbf{x}_\text{worst}$ from  $P$
	\State insert $\textbf{z}$ in $P$
	\State find the new $\textbf{x}_\text{worst}$
	\EndIf
\EndWhile
\State $\textbf{return}$ $\textbf{x}_\text{best}$ and $f(\textbf{x}_\text{best})$
\end{algorithmic}
\end{algorithm}

It is interesting to notice that this algorithm adopts the $(\lambda+1)$ evolutionary scheme, where $\lambda=N$. In fact, at each iteration, one new point is generated (by perturbation and local minimization)
and kept in the population 
if and only if it is better than the worst solution in the current population.

This scheme is extended in two ways. First, when the minimum $\textbf{z}$ enters the population, the point $\textbf{x}$ used in the next iteration will be exactly $\textbf{z}$.
The other improvement is to use a form of restart in order to prevent stagnation and premature convergence.
At each iteration, the algorithm checks whether all the solutions in the population have the same function value. If this condition is met, then $2/3$ of the solutions
in the population are re-initialized, i.e., they are replaced by randomly generated solutions in the function domain, which are subjected to the local search procedure.
Finally, it is interesting to note that, for $N=1$, BHPOP reduces to the standard BH described in the previous section.

For the sake of comparison, the pseudocode provided in Algorithm~\ref{alg:locatelli} describes the population-based approach introduced by Grosso et al. in~\cite{grosso2007} and named PBH.
The distance $d$ mentioned in pseudocode is a kind of parameter of the algorithm and, in~\cite{grosso2007}, is defined as $d(x,y)=|f(x)-f(y)|$ for the case of general benchmark functions (this is also the definition we used in our implementation of PBH).
Note that the pseudocode here reported is simplified with respect to the original code present in \cite{grosso2007} because the parameter $d_{cut}$ is set to $\infty$, as suggested by the authors.


\begin{algorithm}
\caption{PBH by Grosso et al.}
\label{alg:locatelli}
\begin{algorithmic}[0]
\State randomly initialize the $N$ solutions $\{\textbf{x}\}_{1}^{N}$ in the population $P$
\State locally minimize all the solutions $\{\textbf{x}\}_{1}^{N}$ in $P$
\While{termination condition is false}
        \For{$i\gets 1 \mbox{ to } N$}
            \State $\textbf{y}_i \leftarrow$  perturb$(\textbf{x}_i)$
            \State $\textbf{z}_i \leftarrow$    minimize$(f(\textbf{y}_i))$
        \EndFor
        \For{$i\gets 1 \mbox{ to } N$}
            \State $j\gets $ argmin$_k$ $d(\textbf{x}_k, \textbf{z}_i)$
	       \If{$f(\textbf{z}_i) < f(\textbf{x}_j)$}
	           \State replace $\textbf{x}_j$ with $\textbf{z}_i$ in $P$
	       \EndIf
        \EndFor
\EndWhile
\State $\textbf{return}$ $\textbf{x}_\text{best}$ and $f(\textbf{x}_\text{best})$
\end{algorithmic}
\end{algorithm}

The main difference with respect to BHPOP consists in the way the population is managed: PBH uses a generational approach, while BHPOP uses a steady-state mechanism.
The other difference is that BHPOP has a completely elitist behaviour, by maintaining at every iteration the best $N$ individuals so far, while this is not case in PBH, where diversity is ensured by making individuals compete only with those that are locally close. Instead, in order to maintain diversity in the individuals pool, BHPOP relies on the restart mechanism, as described above.

\subsection{Differential Evolution}
\label{dev}

Differential Evolution (DE) was proposed by Storn and Price~\cite{storn1997} as a population-based metaheuristic for optimizing functions. In DE, each individual in the population, denoted by $\textbf{x}$, is subjected to a variation process that involves the recombination of three randomly selected and distinct individuals, namely $\textbf{a}$, $\textbf{b}$, and $\textbf{c}$. This variation results in the creation of a new solution, denoted by $\textbf{z}$, according to the equation $\textbf{z}=\textbf{a} + F\cdot (\textbf{b}-\textbf{c})$, where $F$ is a weight parameter commonly selected from the interval $(0,2]$. A random direction $j$ is then chosen in the $D$-dimensional space, and a new candidate individual, denoted by $\textbf{x}'$, is formed through the use of binomial crossover between $\textbf{x}$ and $\textbf{z}$, with a probability of crossover $p_{cr}$.

\begin{equation}
\textbf{x}'_i=\left\{ 
   \begin{array}{cl}
 \textbf{z}_i & \mbox{if $i=j$ or with probability $p_{cr}$}, \\
{\textbf{x}_i} & \mbox{otherwise}\\
    \end{array}
\right.
\end{equation}

\noindent Finally, assuming function minimization, the new solution $\textbf{x}'$ replaces $\textbf{x}$ if  $f(\textbf{x}') \le f(\textbf{x})$.

\begin{algorithm}
\caption{Differential Evolution}
\label{de}
\begin{algorithmic}[0]
\State initialize the individuals in the population $P$ at random
\While{termination condition is false}
	\For {each individual $\textbf{x} \in P$}
	\State select at random three different individuals $ \textbf{a,b,c} \ne \textbf{x}$ 
	\State combine $\textbf{a,b}$, and $\textbf{c}$ to produce $\textbf{z}$
	\State produce candidate solution $\textbf{x}^{'}$ by crossover between $\textbf{x}$ and $\textbf{z}$
	\If{$f(\textbf{x}') \le f(\textbf{x})$}
	\State replace $\textbf{x}$ with $\textbf{x}'$ in $P$
	\Else 
	\State keep solution $\textbf{x}$ in $P$
	\EndIf
\EndFor
\EndWhile
\State $\textbf{return}$ best solution
\end{algorithmic}
\end{algorithm}

Algorithm~\ref{de} describes the basic DE. There exist more advanced versions which differ
in the intermediate design of $\textbf{z}$ or the crossover, and some hybrids also incorporate
local search.

\subsection{Covariance Matrix Adaptation Evolution Strategy}
\label{cma}

The \textit{Covariance Matrix Adaptation Evolution Strategy} (CMA-ES) derives from Evolution
Strategies but it is a more modern and complex method than the original versions of ES.
CMA-ES evolves the mean value and the covariance matrix of a
multivariate Gaussian distribution $p({\bf x})$ used to draw new solutions in the search space:

$$
  p({\bf x}) = \sqrt{\frac{\det{\bf C}^{-1}}{(2\pi)^l}}
                \exp\left(-\frac{1}{2}(\bf{ x - \bar x})^T{\bf C}^{-1}(\bf {x - \bar x})\right),
$$

\noindent where $\bf{\bar x}$ is the mean and ${\bf C} = (c_{ij})$  is the covariance matrix of $p({\bf x})$. In this matrix, the
$c_{ii}$ are the variances $\sigma_{i}^{2}$ and the off-diagonal elements $c_{ij}$, with $i \ne j$, represent the
covariances. 
The $D$ variances and $D(D-1)/2$ covariances (the matrix is symmetric) needed for parameter evolution
are drawn from this general distribution. It is succinctly written as ${\bf X} \sim {\cal N}({\bf \bar x, C}$).
In CMA-ES a global step size is determined for all parameters and a new vector is obtained by mutation. After generating
and evaluating an offspring population of size $\lambda$, obtained by mutation, the $\mu$ best best individuals among 
the offspring are selected and undergo weighted recombination
which implements a multi-parent arithmetic crossover operator, while a more computationally expensive procedure is required to update
the covariance matrix.
There are a few slightly different versions of CMA-ES in use and the details, which are highly technical, would require a long
discussion which cannot be afforded here.
For this reason, instead of giving the pseudocode for a particular version, we prefer to refer the reader to specialised literature for a more complete description~\cite{hansen1996,back2013}. Arguably, CMA-ES is the most sophisticated among the metaheuristics described herein.
In this work we have used the version provided by the \textit{Nevergrad} library (see Section~\ref{sec:expsetup_alg} below).

\subsection{Particle Swarm Optimization}
\label{pso}

Eberhart and Kennedy~\cite{kennedy1995} proposed a method for optimization called \textit{Particle Swarm Optimization}, which is inspired by animal behavior such as flocks of birds or swarms of insects. The aim of this approach is to simultaneously explore a problem search space by means of a number of ``particles'' with the goal of finding the globally optimum configuration. Here we describe the basic version of the algorithm. PSO works with a population in which each particle $i$, with function value $f(\textbf{x}_i)$, corresponds to a possible solution to the problem. The particles move according to their velocities $\textbf{v}_i$ within the specified continuous search space. At each iteration $\textbf{x}_i^{best}(t)$ and $\textbf{B}(t)$ are computed and updated; $\textbf{x}_i^{best}(t)$ is the best fitness point visited by particle $i$ since the beginning of the search; $\textbf{B}(t)$ is the best function 
value found within the entire population up to time step $t$ and is called \textit{global-best}.

The PSO algorithm incorporates three factors to determine the particles' movement. Firstly, the inertia term maintains the particles on their current path. Secondly, the particles are drawn towards the global best, $\textbf{B}(t)$. And lastly, they are also attracted to their particle-best fitness point, $\textbf{x}_i^{best}(t)$.
The particle's motion from iteration $i$ to $i+1$ is given by the following equations
for the particle's new velocity $\textbf{v}_i(t+1)$ and new position $\textbf{x}_i(t+1)$:
\begin{eqnarray*}
  \textbf{v}_i(t+1)&=&\omega \textbf{v}_i(t) + c_1 r_1(t+1)[\textbf{x}_i^{best}(t)-\textbf{x}_i(t)]
\nonumber\\
                    &&  + \: c_2 r_2(t+1)[\textbf{B}(t)-\textbf{x}_i(t)]\nonumber\\
  \textbf{x}_i(t+1)&=&\textbf{x}_i(t) + \textbf{v}_i(t+1)
\end{eqnarray*}

\noindent where $\omega$, $c_1$ and $c_2$ are scalar parameters to be specified, and $r_1$
and $r_2$ are uniform random variables in the interval $[0,1]$. 
During the initialization phase of the algorithm, the particles are uniformly distributed across the search domain and are assigned zero initial velocity. The algorithm then enters a loop that continues until a specific termination condition is met. During each iteration, $N$ candidate solutions are produced, one for each particle, and the collection of these solutions is utilized to construct the succeeding generation using the dynamical update equations mentioned earlier.
Here, as in the case of DE, the algorithm
described is a bare-bones PSO. Many variants exists but we cannot go into details.

\section{Experimental Setup}
\label{sec:expsetup}

In order to compare the effectiveness of the previously described algorithms, we conducted a series of experiments.
Their setup is detailed hereafter: Section~\ref{sec:expsetup_alg} describes the implementations adopted and the parameter settings for the considered algorithms, Section~\ref{sec:expsetup_ben} describes the benchmark suite on which the 
experiment
was carried out, while Section~\ref{sec:expsetup_des} provides a description of how the 
experiment
was designed.

\subsection{Implementations of the algorithms}
\label{sec:expsetup_alg}


The six metaheuristics described in Section~\ref{sec:mh} were considered, namely: BH and its two population-based variants BHPOP and PBH, DE, PSO, and CMA-ES.

As widely known, any of these algorithms has been proposed in a myriad of variants.
In this experiment we intended to provide a comparison of out-of-the-box implementations of the mentioned metaheuristics.
Therefore, with this aim, we selected the recently proposed \mbox{\textit{Nevergrad}} software library~\cite{nevergrad} and we considered the standard implementations and parametrizations of DE, PSO and CMA-ES as provided by the following Nevergrad classes: \texttt{DE}, \mbox{\texttt{RealSpacePSO}}, and \texttt{CMA}.

For the sake of completeness, we provide the out-of-the-box parametrization adopted in Nevergrad.
DE: the crossover probability is set to $p_{cr}=0.5$, the scale factor is set to $F=0.8$, the mutation operator is ``curr-to-best'', and the population size is set to 30.
PSO: the inertia, cognitive and social weights are set to, respectively, $\omega=\frac{1}{2\log(2)}$, $c_1=c_2=0.5+\log(2)$, while the population size is set to 40.
CMA-ES: the sample size is set to $D^2 /2 + D/2 +3$, while the other settings are as described in~\cite{hansen1996} and~\cite{back2013}.

Since Basin Hopping is not in the collection of Nevergrad's algorithms, we implemented it by considering the local minimizer called \mbox{L-BFGS-B}~\cite{shanno} as provided in the widely adopted \textit{SciPy} software library~\cite{scipy} (in the function called \mbox{\texttt{fmin\_l\_bfgs\_b}}).
In order not to break the black-box assumption, \mbox{L-BFGS-B} is run without providing the gradient function, which L-BFGS-B internally approximates by means of the \textit{finite differences} technique (as implemented in SciPy).
This approximation costs additional objective function evaluations which are accounted for in the budget of evaluations allowed for any BH execution.
Furthermore, BH adopts the standard sharp acceptance criterion and a random perturbation strength sampled from the interval centered in the origin and whose length is $\frac{1}{10}$ of the feasible region range in any dimension\footnote{As described in Section~\ref{sec:expsetup_ben}, the considered search space domain is $[-5,+5]^D$, therefore the perturbation domain of BH, blue BHPOP and PBH is $[-0.5,+0.5]^D$.}.

BHPOP is built on the basis of BH and its only extra parameter is the population size which has been set, on the basis of the problem dimension ~$D$, to $\max\{10,D\}$.
PBH is implemented by faithfully following the description found in \cite{grosso2007} and, for the sake of fairness, we adopted the same population size and local optimizer as used in BHPOP.
Moreover, in order to have a good initial diversity of the population, the initialization is performed by using a high discrepancy generator of vectors.
In particular, we adopted the implementation of the Scrambled Hammersley technique~\cite{faure2010l2} available in the Nevergrad library.

Finally, all the considered algorithms adopt the clipping technique whenever a generated solution violates its box constraint.
In other words, when modifying the coordinates of a solution generates a point that falls outside the allowed coordinate range for some
coordinates, the
point is brought back into the interval by changing the corresponding coordinates to their maximum or minimum value.

\subsection{Benchmark functions}
\label{sec:expsetup_ben}

To put benchmarking practices into perspective it is useful to remember that
it has been proved that the performance of any black box optimization method when it is averaged over all possible discrete functions is the same, as stated in the no free lunch theorems (NFLT)~\cite{wolpert1997no}.
In the continuous case a similar result also holds (see e.g.~\cite{alabert2015}).
As a consequence, the comparison of two algorithms on a given function set of necessarily limited size, would seem
worthless. In fact, even if algorithm $A_1$ is more efficient on the functions of the set, among the possible problems, according to the NFLT,
there always exist other functions on which algorithm $A_2$ is superior.
However, among the possible functions many are random and  do not appear in real problems.
Thus, it is still useful to benchmark an algorithm on a test set which contains functions that are representative of problems that do appear in real-world applications.
In any case, it remains true that results cannot be generalized to other functions that do not belong to the
tested set.  

With this in mind, to compare the performances of the algorithms described in the previous section in a meaningful way, we selected the widely known real-parameter optimization benchmark suite called BBOB~\cite{hansen2009real}.
In particular, we have used the software implementation of BBOB provided in the widespread testing environment \textit{IOH Profiler}~\cite{doerr2018iohprofiler}.

The BBOB benchmark test suite contains $24$ scalable real and single objective test functions which are
described in detail in the report~\cite{hansen2009real}.
The aim in designing the benchmark was to expose most of the typical difficulties encountered in optimization practice.
The function features that were considered important are separability, conditioning, multi-modality, and deception
and the set contains several functions of each of these kinds. Various transformations, including random shifting of the
global optimum, as well as linear and non-linear transformations of the search space were used to generate the desired
features.

The functions $\{f_1, f_2, \ldots, f_{24} \}$ are divided into five groups, each one representative of the corresponding function feature:  separable functions ($f_1$ to $f_5$), functions with low or moderate conditioning ($f_6$ to $f_9$),
unimodal functions with high conditioning ($f_{10}$ to $f_{14}$),
multi-modal functions with adequate global structure ($f_{15}$ to $f_{19}$), and
multi-modal functions with weak global structure ($f_{20}$ to $f_{24}$).

The global minima are sought in the search domain defined by the closed compact $[-5,+5]^D$, where $D$ is the search space dimension which can be set to any positive integer $D \geq 2$.
In the following, we will use the term ``problem'' to refer to a function/dimension pair.

A BBOB problem relies on a basic benchmark function (such as, just to name a few, the well known Rosenbrock, Rastrigin and Schwefel functions) and it may be instantiated in a virtually infinite number of instances by applying linear and non-linear transformations both in the search and the objective spaces.
Therefore, different instances of the same problem may have different global minima.
The IOH Profiler environment allows to track both the transformed and untransformed objective values, together with the difference with respect to the known global minimum.
This is useful in order to correctly aggregate the results over multiple instances, multiple functions, or multiple problems.

\subsection{Design of the experiments}
\label{sec:expsetup_des}

The performances of the algorithms have been compared on the suite of 24 benchmark functions described in the previous section.

Four dimensions $D$ are considered as $D \in \{ 5,10,20,40 \}$, thus our benchmark suite is formed by a total of $24 \times 4 = 96$ problems.
Moreover, for each benchmark problem, $15$ different instances were generated.
Each algorithm has been executed $15$ times per instance, thus the number of executions of a given algorithm on a given problem is $15 \times 15 = 225$.
Overall, by considering all the algorithms and the entire benchmark suite with all the dimensions, our 
experiment
consisted in $6 \times 24 \times 4 \times 225 = 129\,600$ executions.
This large number of executions allows to perform a solid statistical validation of the results.

Any single execution terminates when either a predefined budget of evaluations $\mathit{cap}$ is exhausted or when the difference between the best-so-far value and the known global minimum value is less or equal to a predefined precision error $\mathit{err}$.
We considered the settings $\mathit{cap}=200\,000$ and $\mathit{err}=10^{-8}$ which are somehow standard in the literature and, since IOH Profiler records the trajectory of the best-so-far value in any single execution, they are large/small enough to allow analyses with smaller evaluation budgets or larger precision errors.
Finally, note that, in order to allow a fair comparison, all the recorded errors smaller than $\mathit{err}$ are clipped to $\mathit{err}$.

\section{Experimental Results}
\label{sec:expres}


The trajectories recorded by IOH Profiler for all the algorithms in all the benchmark problems are analyzed under two different perspectives: in the \textit{fixed budget analysis} we investigate how effective are the best objective values obtained by the algorithms within a given amount of evaluations, while in the \textit{fixed target analysis} the emphasis is on the number of evaluations required to reach a given target objective value.

\subsection{Preliminary comparison between BHPOP and PBH}
\label{sec:expres_prel}

Before proceeding with the full scale experiments, because of the similarity between BHPOP and PBH, we wished to compare their performances on the benchmark suite.

For that purpose, we provide in Table~\ref{tab:prel} an experimental comparison between our BHPOP and PBH.
For each dimension $D \in \{5,10,20,40\}$ and for each function group as described in Section~\ref{sec:expsetup_ben}, the table shows the results of the comparison using the notation ``$x/y/z$'' with the following meaning: $x$ is the number of functions where BHPOP significantly outperformed PBH, $y$ is the number of functions where no significant difference is observed, $z$ is the number of functions where BHPOP is significantly outperformed by PBH.
The significance is calculated according to the Mann~Whitney~U test~\cite{stat} with a significance level of $\alpha=0.05$.
The ``Overall'' line provides a summary of the same data throughout the entire benchmark suite, while the Wilcoxon test~\cite{stat} was run using the median results obtained for each benchmark function.

\begin{table}[!h]
    \centering
\begin{tabular}{lp{0.2cm}cp{0.2cm}cp{0.2cm}cp{0.2cm}c}
    \toprule
    \textbf{Function Group} & & $\bm{D=5}$ & & $\bm{D=10}$ & & $\bm{D=20}$ & & $\bm{D=40}$ \\
    \midrule
    Separable           & & 1/4/0 & & 2/3/0 & & 2/3/0 & & 1/4/0 \\
    Low. Cond.          & & 1/3/0 & & 1/3/0 & & 1/3/0 & & 1/3/0 \\
    High Cond.          & & 1/4/0 & & 1/2/2 & & 1/4/0 & & 0/5/0 \\
    Global Structure    & & 5/0/0 & & 5/0/0 & & 4/1/0 & & 1/4/0 \\
    Weak Structure      & & 1/1/3 & & 1/4/0 & & 3/2/0 & & 2/3/0 \\
    \midrule
    Overall             & & 9/12/3 & & 10/12/2 & & 11/13/0 & & 5/19/0 \\
    Wilcoxon p-value    & & 0.811 & & \textbf{0.005} & & \textbf{0.016} & & 0.106 \\
    \bottomrule
\end{tabular}
    \caption{Comparison between BHPOP and PBH. For each dimension (columns) and function group (rows), the notation ``$x/y/z$'' means: BHPOP significantly outperformed PBH in $x$ functions, BHPOP and PBH have no significant difference in $y$ functions, BHPOP is significantly outperformed by PBH in $z$ functions. The significance is according to the Mann~Whitney~U test~\cite{stat} with a significance level of $\alpha=0.05$. The ``Overall'' line provides a summary of the same data throughout the entire benchmark suite, while the Wilcoxon test~\cite{stat} was run using the median results obtained for each benchmark function.}
    \label{tab:prel}
\end{table}

Table~\ref{tab:prel} shows that BHPOP performs slightly better than PBH.
Overall, out of 96 function/dimension pairs, there are 35 successes for BHPOP, 56 ties, and only 5 successes for PBH.
Furthermore, PBH does not exhibit a significant advantage over BHPOP in any of the larger benchmark functions ($D=20,40$).

Based on these results, we will only consider BHPOP as population-based variant of BH in the following experimental comparisons.

\subsection{Fixed Budget Analysis}
\label{sec:expres_budget}

In order to carry out a fair comparison and aggregation of the results among different instances and different problems, the error value $v$ produced by each execution (within a given budget of evaluations) is transformed to the $\mathit{logscore}$ measure defined as $\mathit{logscore}(v) = \log\left(\nicefrac{v}{\mathit{best}}\right)$, where $\mathit{best}$ is the best error observed in all the executions of every algorithm in the same problem instance (within a given budget of evaluations).
The logscores allow for a relative comparison among the algorithms considered in the 
experiment.
Moreover, they can be averaged across multiple instances and problems.
The ideal average logscore of $0$ is obtained by an algorithm which reached the best objective value in all its executions, while the value increases for less effective and less robust algorithms.
Therefore, throughout this section, we will use logscores and average logscores in order to compare the algorithms.

\begin{figure}[!h]
    \centering
    \includegraphics[width=1.0\textwidth]{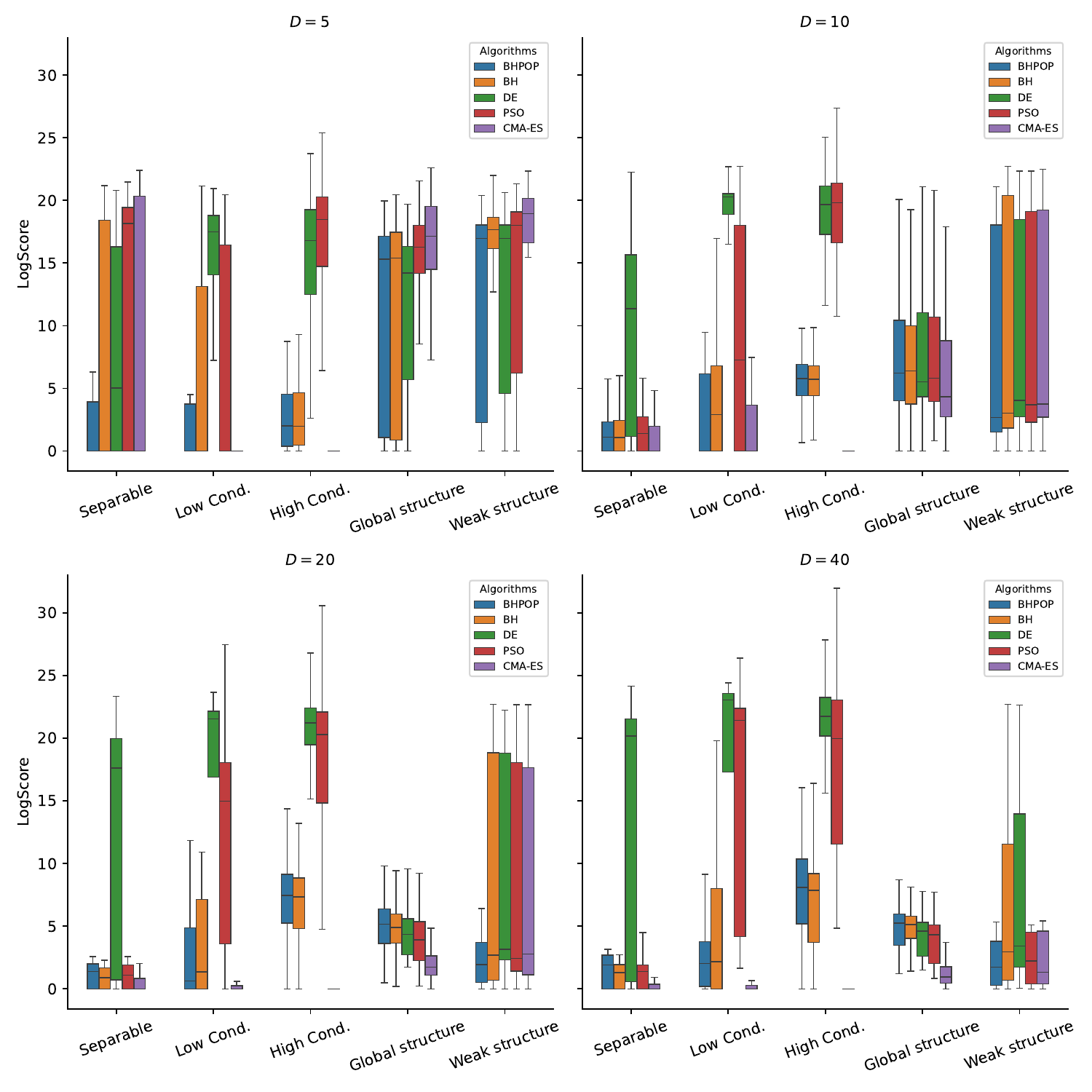}
    \caption{Boxplots of the logscores obtained by the five algorithms, grouped by dimension $D \in \{5,10,20,40\}$ and benchmark function group, considering a budget of $200\,000$ evaluations.}
    \label{fig:logscore200k}
\end{figure}

For each dimension $D \in \{ 5,10,20,40 \}$, Figure~\ref{fig:logscore200k} shows a boxplot graph which is formed by five groups --~one for each group of benchmark functions (see Section~\ref{sec:expsetup_ben})~-- and each group contains a box for each compared algorithm.
The box reflects the distribution of the logscores obtained after $200\,000$ evaluations by an algorithm in all its executions related to the considered group of functions with the considered dimension.

Figure~\ref{fig:logscore200k} can be commented as follows:
\begin{itemize}
    \item regarding the ``BHPOP vs BH'' comparison, it looks that the population of BHPOP, which allows more ``diverse'' attempts of the Basin Hopping search, makes the algorithm more robust (see e.g. the BHPOP and BH boxes in low conditioned functions) and more effective when the landscape is formed by multiple basins of attractions without a regular structure (see the BHPOP and BH boxes in the last group of functions);
    \item when $D=5$, BHPOP and BH are clearly more effective than the competitors in all function groups except for the group of highly conditioned functions \mbox{($f_{10}$ -- $f_{14}$)}, where CMA-ES obtained better results;
    \item CMA-ES starts to outperform the other algorithms from dimension $D=10$ but, notably, BHPOP is competitive with (and sometimes better than) CMA-ES on the supposedly most difficult functions group, i.e. the multi-modal functions with weak global structure ($f_{20}$ -- $f_{24}$);
    \item PSO and DE are, in general, less effective than the other algorithms, though being competitive in the multi-modal functions with adequate global structure ($f_{15}$ -- $f_{19}$).
\end{itemize}

For the sake of completeness, Table~\ref{tab:logscore200k} provides the average logscores obtained --~after $200\,000$ evaluations~-- by each algorithm in all the considered problems, together with the indications of a statistical analysis performed by considering BHPOP as reference algorithm.
In fact, the average logscores of BH, DE, PSO and CMA-ES are marked with $\blacktriangle$ or $\triangledown$ when BHPOP is, respectively, significantly better or significantly worse, according to the Mann~Whitney~U test~\cite{stat} and a significance level of $\alpha=0.05$.
The last line of the table shows the average logscores, aggregated over all the problems with the same dimension.
In this case, the statistical indications are provided according to the Wilcoxon test~\cite{stat} carried out on the $24$ paired average logscores.

\begin{sidewaystable}
    \centering
    \resizebox{1.0\textwidth}{!}{
        \begin{tabular}{cp{0.2cm}rrrrrp{0.2cm}rrrrrp{0.2cm}rrrrrp{0.2cm}rrrrr}
\toprule
 &  & \multicolumn{5}{l}{$\bm{D=5}$} &  & \multicolumn{5}{l}{$\bm{D=10}$} &  & \multicolumn{5}{l}{$\bm{D=20}$} &  & \multicolumn{5}{l}{$\bm{D=40}$} \\
\vspace{-0.2cm} \\
\textbf{Function}  & &  \multicolumn{1}{c}{\textbf{BHPOP}} &           \multicolumn{1}{c}{\textbf{BH}} &           \multicolumn{1}{c}{\textbf{DE}} &          \multicolumn{1}{c}{\textbf{PSO}} &        \multicolumn{1}{c}{\textbf{CMA-ES}}  & &  \multicolumn{1}{c}{\textbf{BHPOP}} &           \multicolumn{1}{c}{\textbf{BH}} &            \multicolumn{1}{c}{\textbf{DE}} &          \multicolumn{1}{c}{\textbf{PSO}} &       \multicolumn{1}{c}{\textbf{CMA-ES}}  & &  \multicolumn{1}{c}{\textbf{BHPOP}} &           \multicolumn{1}{c}{\textbf{BH}} & \multicolumn{1}{c}{\textbf{DE}} &          \multicolumn{1}{c}{\textbf{PSO}} &       \multicolumn{1}{c}{\textbf{CMA-ES}}  & &  \multicolumn{1}{c}{\textbf{BHPOP}} &           \multicolumn{1}{c}{\textbf{BH}} & \multicolumn{1}{c}{\textbf{DE}} & \multicolumn{1}{c}{\textbf{PSO}} &       \multicolumn{1}{c}{\textbf{CMA-ES}} \\
\cmidrule(l){1-1} \cmidrule(l){3-7} \cmidrule(l){9-13} \cmidrule(l){15-19} \cmidrule(l){21-25} 
$f_{1}$           & &                     \textbf{0.00} &  \textbf{0.00} \phantom{$\blacktriangle$} &                     3.11 $\blacktriangle$ &  \textbf{0.00} \phantom{$\blacktriangle$} &   \textbf{0.00} \phantom{$\blacktriangle$}  & &                     \textbf{0.00} &  \textbf{0.00} \phantom{$\blacktriangle$} &                     12.73 $\blacktriangle$ &  \textbf{0.00} \phantom{$\blacktriangle$} &  \textbf{0.00} \phantom{$\blacktriangle$}  & &                     \textbf{0.00} &  \textbf{0.00} \phantom{$\blacktriangle$} &          18.17 $\blacktriangle$ &  \textbf{0.00} \phantom{$\blacktriangle$} &  \textbf{0.00} \phantom{$\blacktriangle$}  & &                     \textbf{0.00} &  \textbf{0.00} \phantom{$\blacktriangle$} &          20.55 $\blacktriangle$ &  0.03 \phantom{$\blacktriangle$} &  \textbf{0.00} \phantom{$\blacktriangle$} \\
$f_{2}$           & &                     \textbf{0.00} &  \textbf{0.00} \phantom{$\blacktriangle$} &                     4.52 $\blacktriangle$ &  \textbf{0.00} \phantom{$\blacktriangle$} &   \textbf{0.00} \phantom{$\blacktriangle$}  & &                              4.95 &           4.88 \phantom{$\blacktriangle$} &                     14.42 $\blacktriangle$ &             \textbf{0.00} $\triangledown$ &             \textbf{0.00} $\triangledown$  & &                             10.77 &          10.66 \phantom{$\blacktriangle$} &          19.41 $\blacktriangle$ &             \textbf{0.00} $\triangledown$ &             \textbf{0.00} $\triangledown$  & &                             12.18 &          12.38 \phantom{$\blacktriangle$} &          21.38 $\blacktriangle$ &             0.05 $\triangledown$ &             \textbf{0.00} $\triangledown$ \\
$f_{3}$           & &                     \textbf{6.66} &                    10.16 $\blacktriangle$ &                     9.70 $\blacktriangle$ &                    16.04 $\blacktriangle$ &                     20.43 $\blacktriangle$  & &                              1.79 &           1.88 \phantom{$\blacktriangle$} &              \textbf{1.39} $\triangledown$ &                     2.35 $\blacktriangle$ &                     2.65 $\blacktriangle$  & &                              1.39 &                      0.94 $\triangledown$ &   \textbf{0.65} $\triangledown$ &                      1.32 $\triangledown$ &                      0.86 $\triangledown$  & &                              1.90 &                      1.34 $\triangledown$ &            0.48 $\triangledown$ &             1.47 $\triangledown$ &             \textbf{0.40} $\triangledown$ \\
$f_{4}$           & &                             16.19 &          16.63 \phantom{$\blacktriangle$} &            \textbf{15.34} $\triangledown$ &                    18.49 $\blacktriangle$ &                     20.02 $\blacktriangle$  & &                              1.35 &           1.40 \phantom{$\blacktriangle$} &              \textbf{0.71} $\triangledown$ &                     1.56 $\blacktriangle$ &                     1.86 $\blacktriangle$  & &                              1.81 &                      1.40 $\triangledown$ &   \textbf{0.54} $\triangledown$ &                      1.51 $\triangledown$ &                      1.05 $\triangledown$  & &                              2.57 &                      1.77 $\triangledown$ &   \textbf{0.50} $\triangledown$ &             1.66 $\triangledown$ &                      0.53 $\triangledown$ \\
$f_{5}$           & &                     \textbf{0.00} &  \textbf{0.00} \phantom{$\blacktriangle$} &                     4.84 $\blacktriangle$ &                    19.24 $\blacktriangle$ &   \textbf{0.00} \phantom{$\blacktriangle$}  & &                     \textbf{0.00} &  \textbf{0.00} \phantom{$\blacktriangle$} &                     16.37 $\blacktriangle$ &                    20.87 $\blacktriangle$ &  \textbf{0.00} \phantom{$\blacktriangle$}  & &                     \textbf{0.00} &  \textbf{0.00} \phantom{$\blacktriangle$} &          20.19 $\blacktriangle$ &                    22.25 $\blacktriangle$ &  \textbf{0.00} \phantom{$\blacktriangle$}  & &                     \textbf{0.00} &  \textbf{0.00} \phantom{$\blacktriangle$} &          21.51 $\blacktriangle$ &           23.42 $\blacktriangle$ &  \textbf{0.00} \phantom{$\blacktriangle$} \\
\cmidrule(l){1-1} \cmidrule(l){3-7} \cmidrule(l){9-13} \cmidrule(l){15-19} \cmidrule(l){21-25} 
$f_{6}$           & &                              0.89 &           0.75 \phantom{$\blacktriangle$} &                    16.38 $\blacktriangle$ &             \textbf{0.00} $\triangledown$ &              \textbf{0.00} $\triangledown$  & &                              5.43 &           5.26 \phantom{$\blacktriangle$} &                     20.12 $\blacktriangle$ &                      0.77 $\triangledown$ &             \textbf{0.00} $\triangledown$  & &                              8.25 &                      7.89 $\triangledown$ &          22.19 $\blacktriangle$ &                    16.74 $\blacktriangle$ &             \textbf{0.00} $\triangledown$  & &                              9.40 &           9.20 \phantom{$\blacktriangle$} &          23.53 $\blacktriangle$ &           23.44 $\blacktriangle$ &             \textbf{0.00} $\triangledown$ \\
$f_{7}$           & &                             14.45 &                    15.41 $\blacktriangle$ &             \textbf{5.14} $\triangledown$ &                     13.87 $\triangledown$ &                      14.17 $\triangledown$  & &                             13.10 &                    13.38 $\blacktriangle$ &  \textbf{12.99} \phantom{$\blacktriangle$} &                    14.55 $\blacktriangle$ &                     13.05 $\triangledown$  & &                              1.51 &                     1.79 $\blacktriangle$ &           2.18 $\blacktriangle$ &                     2.79 $\blacktriangle$ &             \textbf{1.25} $\triangledown$  & &                              2.35 &           2.34 \phantom{$\blacktriangle$} &            2.20 $\triangledown$ &            3.08 $\blacktriangle$ &             \textbf{0.81} $\triangledown$ \\
$f_{8}$           & &                     \textbf{0.00} &                     0.62 $\blacktriangle$ &                    17.96 $\blacktriangle$ &                     3.11 $\blacktriangle$ &                      1.23 $\blacktriangle$  & &                     \textbf{0.00} &                     2.73 $\blacktriangle$ &                     20.39 $\blacktriangle$ &                     6.38 $\blacktriangle$ &                     1.94 $\blacktriangle$  & &                     \textbf{0.09} &                     2.63 $\blacktriangle$ &          21.93 $\blacktriangle$ &                    11.87 $\blacktriangle$ &           1.76 \phantom{$\blacktriangle$}  & &                     \textbf{0.66} &           2.48 \phantom{$\blacktriangle$} &          23.39 $\blacktriangle$ &           18.66 $\blacktriangle$ &                     1.76 $\blacktriangle$ \\
$f_{9}$           & &                     \textbf{0.00} &  \textbf{0.00} \phantom{$\blacktriangle$} &                    18.19 $\blacktriangle$ &                     7.03 $\blacktriangle$ &                      0.35 $\blacktriangle$  & &                     \textbf{0.00} &           0.09 \phantom{$\blacktriangle$} &                     20.37 $\blacktriangle$ &                    15.39 $\blacktriangle$ &                     0.53 $\blacktriangle$  & &                     \textbf{0.04} &                     3.74 $\blacktriangle$ &          21.62 $\blacktriangle$ &                    17.86 $\blacktriangle$ &           0.70 \phantom{$\blacktriangle$}  & &                     \textbf{0.46} &                     3.66 $\blacktriangle$ &          22.83 $\blacktriangle$ &           21.80 $\blacktriangle$ &                     0.88 $\blacktriangle$ \\
\cmidrule(l){1-1} \cmidrule(l){3-7} \cmidrule(l){9-13} \cmidrule(l){15-19} \cmidrule(l){21-25} 
$f_{10}$          & &                              0.68 &           0.74 \phantom{$\blacktriangle$} &                    18.50 $\blacktriangle$ &                    19.52 $\blacktriangle$ &              \textbf{0.00} $\triangledown$  & &                              6.17 &           6.09 \phantom{$\blacktriangle$} &                     22.87 $\blacktriangle$ &                    22.68 $\blacktriangle$ &             \textbf{0.00} $\triangledown$  & &                             11.06 &                     10.77 $\triangledown$ &          26.09 $\blacktriangle$ &                    25.82 $\blacktriangle$ &             \textbf{0.00} $\triangledown$  & &                             12.36 &          12.32 \phantom{$\blacktriangle$} &          27.51 $\blacktriangle$ &           28.57 $\blacktriangle$ &             \textbf{0.00} $\triangledown$ \\
$f_{11}$          & &                              3.97 &           4.04 \phantom{$\blacktriangle$} &                    18.27 $\blacktriangle$ &                    17.59 $\blacktriangle$ &              \textbf{0.00} $\triangledown$  & &                              6.88 &           6.82 \phantom{$\blacktriangle$} &                     20.02 $\blacktriangle$ &                    19.81 $\blacktriangle$ &             \textbf{0.00} $\triangledown$  & &                              8.17 &                      8.01 $\triangledown$ &          21.92 $\blacktriangle$ &                    21.50 $\blacktriangle$ &             \textbf{0.00} $\triangledown$  & &                             10.01 &                      8.65 $\triangledown$ &          23.10 $\blacktriangle$ &           22.70 $\blacktriangle$ &             \textbf{0.00} $\triangledown$ \\
$f_{12}$          & &                              0.68 &           0.69 \phantom{$\blacktriangle$} &                    17.91 $\blacktriangle$ &                    19.13 $\blacktriangle$ &              \textbf{0.00} $\triangledown$  & &                              1.91 &           1.92 \phantom{$\blacktriangle$} &                     18.22 $\blacktriangle$ &                    18.90 $\blacktriangle$ &             \textbf{0.00} $\triangledown$  & &                              2.95 &           2.75 \phantom{$\blacktriangle$} &          20.04 $\blacktriangle$ &                    19.06 $\blacktriangle$ &             \textbf{0.00} $\triangledown$  & &                              4.04 &                      3.02 $\triangledown$ &          21.64 $\blacktriangle$ &           19.52 $\blacktriangle$ &             \textbf{0.00} $\triangledown$ \\
$f_{13}$          & &                              5.76 &           5.89 \phantom{$\blacktriangle$} &                    14.77 $\blacktriangle$ &                    19.31 $\blacktriangle$ &              \textbf{0.01} $\triangledown$  & &                              7.18 &           7.00 \phantom{$\blacktriangle$} &                     19.91 $\blacktriangle$ &                    19.68 $\blacktriangle$ &             \textbf{1.23} $\triangledown$  & &                              6.90 &             \textbf{6.13} $\triangledown$ &          19.97 $\blacktriangle$ &                    17.60 $\blacktriangle$ &                    11.28 $\blacktriangle$  & &                              5.33 &             \textbf{3.06} $\triangledown$ &          16.55 $\blacktriangle$ &           12.79 $\blacktriangle$ &                    10.98 $\blacktriangle$ \\
$f_{14}$          & &                              1.74 &           1.73 \phantom{$\blacktriangle$} &                     9.00 $\blacktriangle$ &                     5.99 $\blacktriangle$ &              \textbf{0.00} $\triangledown$  & &                              4.74 &           4.72 \phantom{$\blacktriangle$} &                     14.41 $\blacktriangle$ &                     7.16 $\blacktriangle$ &             \textbf{0.00} $\triangledown$  & &                              7.08 &           7.09 \phantom{$\blacktriangle$} &          18.72 $\blacktriangle$ &                     8.85 $\blacktriangle$ &             \textbf{0.00} $\triangledown$  & &                              7.94 &           7.92 \phantom{$\blacktriangle$} &          20.59 $\blacktriangle$ &           10.40 $\blacktriangle$ &             \textbf{0.00} $\triangledown$ \\
\cmidrule(l){1-1} \cmidrule(l){3-7} \cmidrule(l){9-13} \cmidrule(l){15-19} \cmidrule(l){21-25} 
$f_{15}$          & &                              0.14 &  \textbf{0.12} \phantom{$\blacktriangle$} &                    17.46 $\blacktriangle$ &                    19.35 $\blacktriangle$ &                     20.12 $\blacktriangle$  & &                             15.77 &            \textbf{14.00} $\triangledown$ &                     19.82 $\blacktriangle$ &                    20.05 $\blacktriangle$ &                    19.82 $\blacktriangle$  & &                              1.77 &             \textbf{0.67} $\triangledown$ &           2.58 $\blacktriangle$ &                     2.27 $\blacktriangle$ &                      1.53 $\triangledown$  & &                              1.98 &                      1.20 $\triangledown$ &            1.90 $\triangledown$ &             1.76 $\triangledown$ &             \textbf{0.47} $\triangledown$ \\
$f_{16}$          & &                             17.37 &                    17.72 $\blacktriangle$ &                     14.61 $\triangledown$ &                     14.71 $\triangledown$ &             \textbf{13.48} $\triangledown$  & &                              9.56 &           9.63 \phantom{$\blacktriangle$} &                       8.98 $\triangledown$ &                      8.22 $\triangledown$ &             \textbf{5.91} $\triangledown$  & &                              5.86 &           5.76 \phantom{$\blacktriangle$} &            5.49 $\triangledown$ &                      4.59 $\triangledown$ &             \textbf{1.84} $\triangledown$  & &                              5.41 &                      5.32 $\triangledown$ &            5.07 $\triangledown$ &             4.49 $\triangledown$ &             \textbf{1.14} $\triangledown$ \\
$f_{17}$          & &                             15.35 &          15.53 \phantom{$\blacktriangle$} &             \textbf{2.38} $\triangledown$ &                     13.21 $\triangledown$ &           15.02 \phantom{$\blacktriangle$}  & &                              6.80 &                     7.40 $\blacktriangle$ &                       5.74 $\triangledown$ &           6.69 \phantom{$\blacktriangle$} &             \textbf{4.42} $\triangledown$  & &                              7.10 &                      6.63 $\triangledown$ &            6.38 $\triangledown$ &                      6.37 $\triangledown$ &             \textbf{2.91} $\triangledown$  & &                              6.56 &                      6.19 $\triangledown$ &            5.79 $\triangledown$ &             5.80 $\triangledown$ &             \textbf{1.79} $\triangledown$ \\
$f_{18}$          & &                             17.18 &                    17.53 $\blacktriangle$ &             \textbf{7.41} $\triangledown$ &                     16.38 $\triangledown$ &           16.96 \phantom{$\blacktriangle$}  & &                              5.66 &                     5.97 $\blacktriangle$ &                       4.27 $\triangledown$ &                      5.03 $\triangledown$ &             \textbf{3.37} $\triangledown$  & &                              5.69 &                      5.14 $\triangledown$ &            4.62 $\triangledown$ &                      4.58 $\triangledown$ &             \textbf{1.94} $\triangledown$  & &                              5.92 &                      5.73 $\triangledown$ &            5.01 $\triangledown$ &             4.96 $\triangledown$ &             \textbf{1.65} $\triangledown$ \\
$f_{19}$          & &                              5.62 &  \textbf{5.14} \phantom{$\blacktriangle$} &                    14.94 $\blacktriangle$ &                    15.45 $\blacktriangle$ &                     15.88 $\blacktriangle$  & &                              1.60 &             \textbf{0.93} $\triangledown$ &                      4.46 $\blacktriangle$ &                     3.46 $\blacktriangle$ &                     3.27 $\blacktriangle$  & &                              3.94 &                     4.09 $\blacktriangle$ &            2.92 $\triangledown$ &                      1.74 $\triangledown$ &             \textbf{1.44} $\triangledown$  & &                              3.72 &                     4.35 $\blacktriangle$ &            2.80 $\triangledown$ &             2.04 $\triangledown$ &             \textbf{0.92} $\triangledown$ \\
\cmidrule(l){1-1} \cmidrule(l){3-7} \cmidrule(l){9-13} \cmidrule(l){15-19} \cmidrule(l){21-25} 
$f_{20}$          & &                             16.84 &                     16.71 $\triangledown$ &            \textbf{10.32} $\triangledown$ &                     16.14 $\triangledown$ &                     18.85 $\blacktriangle$  & &                              2.40 &                     2.62 $\blacktriangle$ &              \textbf{1.92} $\triangledown$ &                     3.21 $\blacktriangle$ &                     3.64 $\blacktriangle$  & &                              0.43 &             \textbf{0.39} $\triangledown$ &           0.70 $\blacktriangle$ &                     0.76 $\blacktriangle$ &                     1.00 $\blacktriangle$  & &                              0.24 &             \textbf{0.20} $\triangledown$ &           0.57 $\blacktriangle$ &  0.23 \phantom{$\blacktriangle$} &                     0.44 $\blacktriangle$ \\
$f_{21}$          & &                             11.51 &                    17.05 $\blacktriangle$ &  \textbf{9.55} \phantom{$\blacktriangle$} &                    14.09 $\blacktriangle$ &                     16.52 $\blacktriangle$  & &                    \textbf{13.20} &                    19.65 $\blacktriangle$ &                     17.81 $\blacktriangle$ &                    17.85 $\blacktriangle$ &                    16.43 $\blacktriangle$  & &                     \textbf{8.92} &                    20.14 $\blacktriangle$ &          20.02 $\blacktriangle$ &                    17.53 $\blacktriangle$ &                    16.69 $\blacktriangle$  & &                     \textbf{8.21} &                    19.85 $\blacktriangle$ &          20.70 $\blacktriangle$ &           16.73 $\blacktriangle$ &                    14.85 $\blacktriangle$ \\
$f_{22}$          & &                     \textbf{9.58} &                    15.41 $\blacktriangle$ &                    13.06 $\blacktriangle$ &                    14.09 $\blacktriangle$ &                     16.62 $\blacktriangle$  & &                    \textbf{15.04} &                    20.48 $\blacktriangle$ &                     18.93 $\blacktriangle$ &                    17.61 $\blacktriangle$ &                    19.96 $\blacktriangle$  & &                    \textbf{10.83} &                    14.44 $\blacktriangle$ &          14.41 $\blacktriangle$ &                    13.64 $\blacktriangle$ &                    14.10 $\blacktriangle$  & &                     \textbf{3.77} &                     5.89 $\blacktriangle$ &           6.79 $\blacktriangle$ &            5.51 $\blacktriangle$ &                     5.58 $\blacktriangle$ \\
$f_{23}$          & &                             13.05 &          13.06 \phantom{$\blacktriangle$} &                    14.02 $\blacktriangle$ &          12.95 \phantom{$\blacktriangle$} &  \textbf{11.34} \phantom{$\blacktriangle$}  & &                              2.30 &           2.25 \phantom{$\blacktriangle$} &                      2.89 $\blacktriangle$ &             \textbf{2.04} $\triangledown$ &                     2.52 $\blacktriangle$  & &                              2.84 &           2.86 \phantom{$\blacktriangle$} &           3.35 $\blacktriangle$ &  \textbf{2.72} \phantom{$\blacktriangle$} &                     3.05 $\blacktriangle$  & &                              3.88 &           3.88 \phantom{$\blacktriangle$} &           4.17 $\blacktriangle$ &  3.76 \phantom{$\blacktriangle$} &             \textbf{3.55} $\triangledown$ \\
$f_{24}$          & &                    \textbf{10.56} &          10.66 \phantom{$\blacktriangle$} &                    14.00 $\blacktriangle$ &                    14.14 $\blacktriangle$ &                     14.36 $\blacktriangle$  & &                              2.18 &             \textbf{1.88} $\triangledown$ &                      3.83 $\blacktriangle$ &                     3.52 $\blacktriangle$ &                     3.32 $\blacktriangle$  & &                              1.80 &             \textbf{0.92} $\triangledown$ &           2.54 $\blacktriangle$ &                     2.03 $\blacktriangle$ &                      1.65 $\triangledown$  & &                              1.79 &                      1.10 $\triangledown$ &           1.86 $\blacktriangle$ &             1.47 $\triangledown$ &             \textbf{0.71} $\triangledown$ \\
\cmidrule(l){1-1} \cmidrule(l){3-7} \cmidrule(l){9-13} \cmidrule(l){15-19} \cmidrule(l){21-25} 
\cmidrule(l){1-1} \cmidrule(l){3-7} \cmidrule(l){9-13} \cmidrule(l){15-19} \cmidrule(l){21-25} 
\textbf{Overall}  & &                     \textbf{7.01} &                     7.73 $\blacktriangle$ &                    12.14 $\blacktriangle$ &                    12.91 $\blacktriangle$ &            8.97 \phantom{$\blacktriangle$}  & &                              5.33 &           5.87 \phantom{$\blacktriangle$} &                     12.65 $\blacktriangle$ &                     9.91 $\blacktriangle$ &  \textbf{4.33} \phantom{$\blacktriangle$}  & &                              4.55 &           5.20 \phantom{$\blacktriangle$} &          12.36 $\blacktriangle$ &                     9.31 $\blacktriangle$ &  \textbf{2.63} \phantom{$\blacktriangle$}  & &                              4.61 &           5.08 \phantom{$\blacktriangle$} &          12.52 $\blacktriangle$ &  9.76 \phantom{$\blacktriangle$} &             \textbf{1.94} $\triangledown$ \\
\bottomrule
\end{tabular}

    }
    \caption{Average logscores obtained by the five algorithms in all the investigated benchmark functions and dimensions, considering a budget of $200\,000$ evaluations. Best results per problem are denoted in boldface. The logscores marked with $\blacktriangle$ and $\triangledown$ are, respectively, significantly worse or better than the logscore obtained by BHPOP, according to the Mann Whitney U test with $\alpha=0.05$. The last row of the table contains the macro-averaged logscores which are marked as before, according to the Wilcoxon test. Horizontal lines are used to group the functions with similar characteristics, as described in Section~\ref{sec:expsetup_ben}.}
    \label{tab:logscore200k}
\end{sidewaystable}

Table~\ref{tab:logscore200k} confirms the general comments made before, but allows to derive indications about the statistical significance and more detailed observations as follows.
Averaging across all the considered benchmark problems, there is only one comparison where BHPOP is significantly worse than a competitor algorithm, i.e. with respect to CMA-ES in the 40-dimensional problems.
Nevertheless, there are six 40-dimensional functions (namely: $f_8$, $f_9$, $f_{13}$, $f_{20}$, $f_{21}$ and $f_{22}$) where BHPOP obtains significantly better results than CMA-ES.
Regarding the comparison ``BHPOP vs BH'', there are several cases where the basic BH is better than its population-based variant.
Anyway, when that happens, the logscores of BH are slightly better than those of BHPOP while, conversely, when BHPOP outperforms BH, its logscores are largely better (see e.g. the results for $f_{21}$ and $f_{22}$).
Finally, as seen before, PSO and DE look competitive in the fourth group of functions \mbox{($f_{15}$ -- $f_{19}$)}.

\begin{figure}[H]
    \centering
    \includegraphics[width=0.8\textwidth]{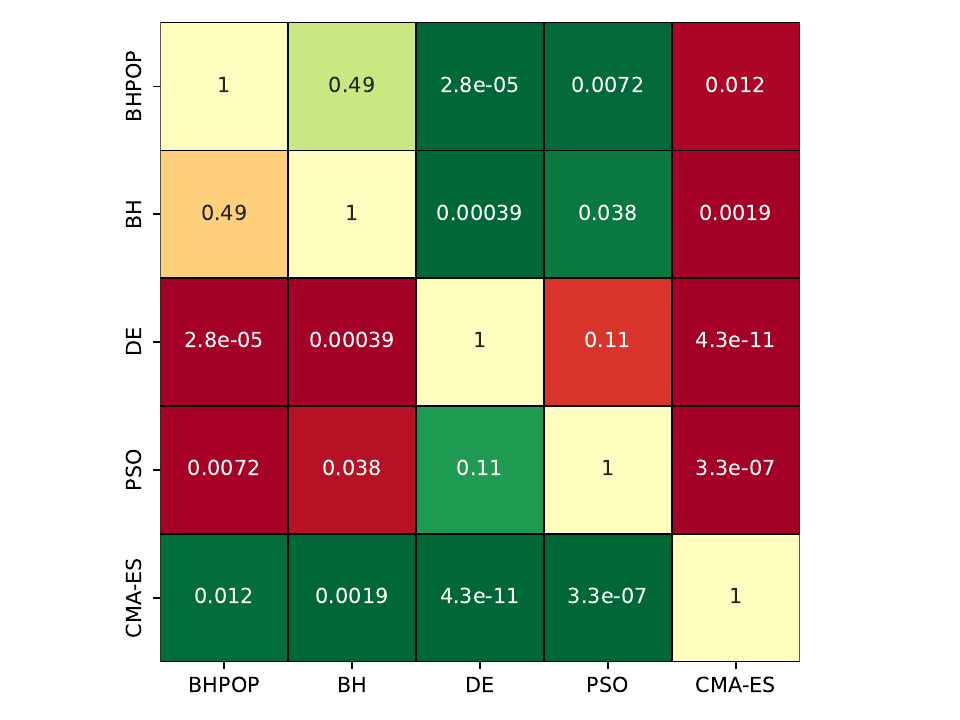}
    \caption{Heatmap showing all the pairwise comparisons among the five algorithms in all the $4 \times 24 = 96$ problems, considering a budget of $200\,000$ evaluations. The entries are greenish or reddish when the row-algorithm is, respectively, better or worse than the column-algorithm. The grade of green or red is set on the basis of the p-value computed according to the Conover post-hoc test with Benjamini-Hochberg adjustment. These \mbox{p-values} are also provided in the heatmap. The omnibus Friedman test rejected the equivalence of effectiveness among the five algorithms with a p-value smaller than $10^{-11}$.}
    \label{fig:heatmap200k}
\end{figure}

In order to have 
synthetic
global indications, we also carried out a statistical analysis on the average logscores obtained by the five algorithms in all the $4 \times 24 = 96$ problems, within the maximum allowed budget of $200\,000$ evaluations. 
The omnibus Friedman test~\cite{stat} rejects the equivalence of effectiveness among the five algorithms with a p-value smaller than $10^{-11}$.
Therefore, a Conover post-hoc test has been carried out by considering the Benjamini-Hochberg adjustment scheme in order to mitigate the statistical family-wise error rate~\cite{hollander2013nonparametric}.
The results of all the pairwise comparisons are provided in the heatmap of Figure~\ref{fig:heatmap200k}.
The entries are greenish or reddish when the row-algorithm is, respectively, better or worse than the column-algorithm.
The grades of green or red are set on the basis of the adjusted Conover p-values, which are also provided in the entries.

Figure~\ref{fig:heatmap200k} shows that a standing of the algorithms may be divided in three bands: CMA-ES in the first band,BHPOP and BH in the second band, and PSO and DE in the third band.
In fact, the more sophisticated scheme of CMA-ES is significantly more effective than the competitors, at least with the
considered budget of evaluations.
However, the p-values of the comparisons ``CMA-ES vs BHPOP'' and ``CMA-ES vs BH'' are several orders of magnitude larger than those of the comparisons ``CMA-ES vs PSO'' and ``CMA-ES vs DE''.

\begin{figure}[H]
    \centering
    \includegraphics[width=1.00\textwidth]{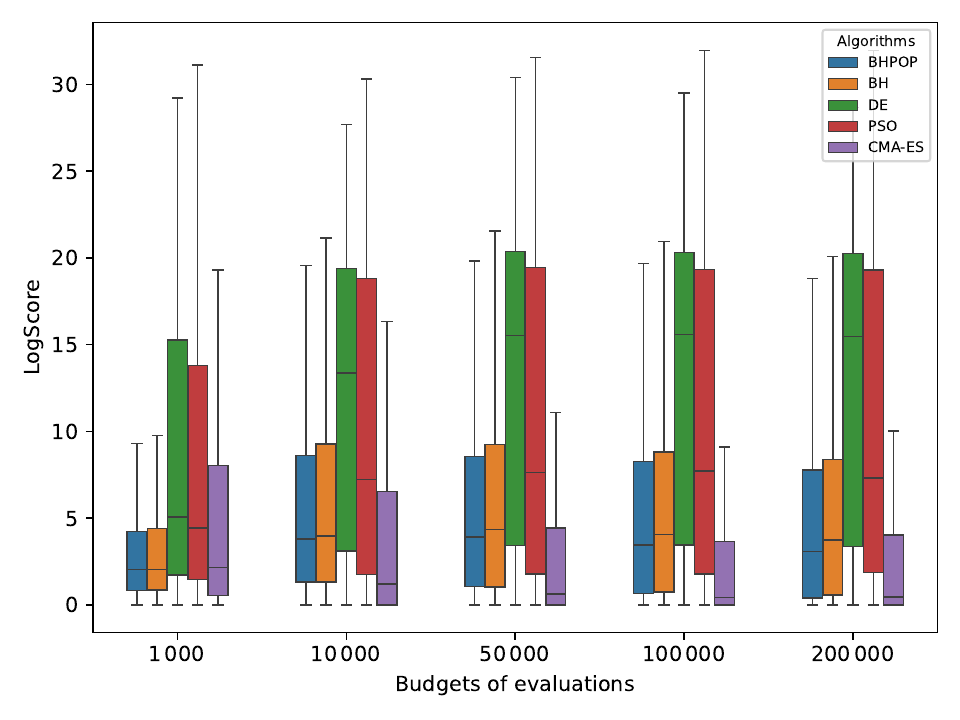}
    \caption{Boxplots of the logscores obtained by the five algorithms in all the executions over all the problem instances, considering different budgets of evaluations.}
    \label{fig:variablebudget}
\end{figure}

Finally, the effectiveness of the algorithms has been compared also by considering lower budgets of evaluations.
In Figure~\ref{fig:variablebudget} we provide the boxplots of the logscores obtained by the algorithms over the entire benchmark suite considering five different budgets of evaluations as follows: $1000$, $10\,000$, $50\,000$, $100\,000$ and $200\,000$.

Interestingly, Figure~\ref{fig:variablebudget} shows that, with a very low budget of $1000$ evaluations, BHPOP and BH are more robust and effective than the other competitors, while the previously made observations seems to be valid for budgets larger or equal to $10\,000$.

\subsection{Fixed Target Analysis}
\label{sec:expres_target}

The aim of the fixed target analysis is to compare the number of evaluations required by the algorithms to reach a given target objective value.
For fairer aggregations and comparisons, since the benchmark functions have different global minima, and by recalling that such global minima are known, here we considered target precision errors, i.e. differences with respect to the known global minimum values.
In particular, we have investigated five target errors as follows: $t \in \{ 10^{-8}, 10^{-4}, 0.01, 0.1, 1 \}$.

We denote with $T(A,f,i,b,t)$ the number of evaluations required by an algorithm $A$, in its $i$-th execution on the problem $f$, to reach a precision error smaller or equal to the target $t$, within the allotted budget of evaluations $b$.
Hence, $T(A,f,i,b,t)$ may assume integer values in $[1,b]$, or $\infty$ when the target is not reached. 
As described in Section~\ref{sec:expsetup_des}, we consider $b=200\,000$ as maximum allowed budget of evaluations.

In order to measure the performances of the algorithms, the following statistics are taken into account:
\begin{equation}
    \mathit{SR} = \frac{1}{r} \sum_{i=1}^r \mathbbm{1}\left( T(A,f,i,b,t)<\infty \right) ,
    \label{eq:sr}
\end{equation}
\begin{equation}
    \mathit{AR} = \frac{1}{r} \sum_{i=1}^r \min \{ T(A,f,i,b,t), b \} ,
    \label{eq:ar}
\end{equation}
\begin{equation}
    \mathit{ERT} = \frac{\mathit{AR}}{\mathit{SR}} ,
    \label{eq:ert}
\end{equation}
where: $r$ is the number of executions of algorithm $A$ on problem $f$, while $\mathbbm{1}(\xi)$ is the indicator function of the event $\xi$.

The empirical success rate $\mathit{SR} \in [0,1]$ is the fraction of executions in which the algorithm reached the given target within the allowed budget of evaluations.
The average runtime $\mathit{AR} \in [1,b]$ is the average number of evaluations required by an algorithm to reach the given target, counting $b$ evaluations for non-successful executions.
The expected runtime $\mathit{ERT}$ penalizes $\mathit{AR}$ by dividing it by the success rate $\mathit{SR}$. Practically, $\mathit{ERT}$ corresponds to the expected number of evaluations required to reach the target $t$ by a multistart version of the algorithm $A$, i.e. an algorithm which restarts its execution after $b$ evaluations and stops as soon as the target $t$ is reached.
It is worth noting that $\mathit{SR}=0$ implies $\mathit{ERT}=\infty$.
These measures are widely used in the literature (see e.g.~\cite{wang2022iohanalyzer}) and can be safely aggregated across multiple instances and multiple problems.

\begin{figure}[!h]
    \centering
    \includegraphics[width=1.00\textwidth]{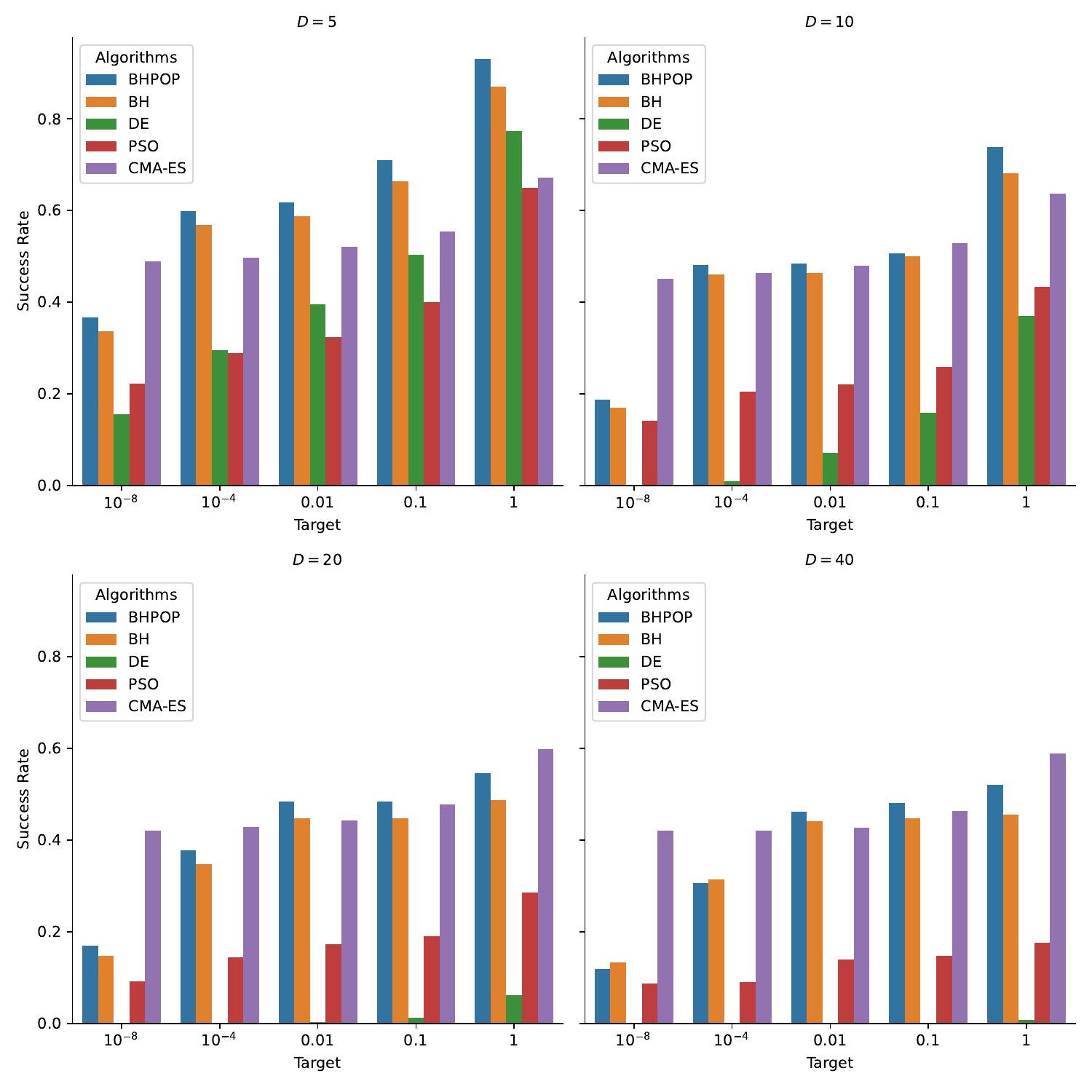}
    \caption{Average success rates achieved by the five algorithms, grouped by dimension and target precision.}
    \label{fig:sr}
\end{figure}

For each dimension $D \in \{ 5,10,20,40 \}$, Figure~\ref{fig:sr} shows a histogram graph formed by five groups --~one for each target~-- with a bar for each algorithm.
The height of the bar reflects the success rate $\mathit{SR}$ of the algorithm aggregated on all the benchmark functions with the given dimension.

Figure~\ref{fig:sr} can be commented as follows:
\begin{itemize}
    \item as expected, the success rates decreases when the dimension increases, though the success rates of CMA-ES seems to be less affected by the increase of dimension;
    \item BHPOP has a slightly larger success rate than BH in almost all the cases;
    \item BHPOP and BH are, blue on average, more successful than PSO and DE;
    \item with a target greater or equal to $10^{-4}$, BHPOP seems to be competitive with CMA-ES and, most notably, with target $t=0.01$ it is always more successful than CMA-ES.
\end{itemize}

To further analyze this last point, Table~\ref{tab:target001} provides the success rates and the expected runtimes for each single benchmark function investigated with target $t=0.01$ and dimension $D=40$.

\begin{table}[!h]
    \centering
    \resizebox{1.00\textwidth}{!}{
        \begin{tabular}{cp{0.2cm}cccccp{0.2cm}rrrrr}
	\toprule
	& & \multicolumn{5}{l}{\textbf{SR}} & & \multicolumn{5}{l}{\textbf{ERT}} \\
	\vspace{-0.2cm} \\
	\textbf{Function} & & \textbf{BHPOP} & \textbf{BH} & \textbf{DE} & \textbf{PSO} & \textbf{CMA-ES} & & \textbf{BHPOP} & \textbf{BH} & \textbf{DE} & \textbf{PSO} & \textbf{CMA-ES} \\
	\cmidrule(l){1-1} \cmidrule(l){3-7} \cmidrule(l){9-13}
	$f_{1}$   & & \textbf{1.00} & \textbf{1.00} & 0.00 & \textbf{1.00} &   \textbf{1.00} & &       \textbf{83} &       85 & $\infty$ &    7919 &    2230 \\
	$f_{2}$   & & 0.91 & 0.97 & 0.00 & \textbf{1.00} &   \textbf{1.00} & &   141533 &   134692 & $\infty$ &   \textbf{24366} &   43542 \\
	$f_{3}$   & & 0.00 & 0.00 & 0.00 & 0.00 &   0.00 & &      $\infty$ &      $\infty$ & $\infty$ &     $\infty$ &     $\infty$ \\
	$f_{4}$   & & 0.00 & 0.00 & 0.00 & 0.00 &   0.00 & &      $\infty$ &      $\infty$ & $\infty$ &     $\infty$ &     $\infty$ \\
	$f_{5}$   & & \textbf{1.00} & \textbf{1.00} & 0.00 & 0.00 &   \textbf{1.00} & &      \textbf{117} &      121 & $\infty$ &     $\infty$ &    4547 \\
	\cmidrule(l){1-1} \cmidrule(l){3-7} \cmidrule(l){9-13}
	$f_{6}$   & & \textbf{1.00} & \textbf{1.00} & 0.00 & 0.00 &   \textbf{1.00} & &    35704 &    35180 & $\infty$ &     $\infty$ &   \textbf{11561} \\
	$f_{7}$   & & 0.00 & 0.00 & 0.00 & 0.00 &   0.00 & &      $\infty$ &      $\infty$ & $\infty$ &     $\infty$ &     $\infty$ \\
	$f_{8}$   & & \textbf{1.00} & 0.89 & 0.00 & 0.15 &   0.91 & &    \textbf{15559} &    32686 & $\infty$ & 1277990 &   77342 \\
	$f_{9}$   & & \textbf{1.00} & 0.82 & 0.00 & 0.00 &   0.96 & &    \textbf{11854} &    54516 & $\infty$ &     $\infty$ &   65501 \\
	\cmidrule(l){1-1} \cmidrule(l){3-7} \cmidrule(l){9-13}
	$f_{10}$  & & 0.86 & 0.89 & 0.00 & 0.00 &   \textbf{1.00} & &   171796 &   168694 & $\infty$ &     $\infty$ &   \textbf{49549} \\
	$f_{11}$  & & \textbf{1.00} & \textbf{1.00} & 0.00 & 0.00 &   \textbf{1.00} & &    12356 &     \textbf{8751} & $\infty$ &     $\infty$ &   24537 \\
	$f_{12}$  & & \textbf{1.00} & \textbf{1.00} & 0.00 & 0.04 &   \textbf{1.00} & &     5674 &     \textbf{5482} & $\infty$ & 4328758 &   21069 \\
	$f_{13}$  & & 0.73 & \textbf{0.98} & 0.00 & 0.05 &   0.11 & &   140351 &    \textbf{55327} & $\infty$ & 3937568 & 1616223 \\
	$f_{14}$  & & \textbf{1.00} & \textbf{1.00} & 0.00 & \textbf{1.00} &   \textbf{1.00} & &     \textbf{1143} &     1148 & $\infty$ &   14816 &    3712 \\
	\cmidrule(l){1-1} \cmidrule(l){3-7} \cmidrule(l){9-13}
	$f_{15}$  & & 0.00 & 0.00 & 0.00 & 0.00 &   0.00 & &      $\infty$ &      $\infty$ & $\infty$ &     $\infty$ &     $\infty$ \\
	$f_{16}$  & & 0.00 & 0.00 & 0.00 & 0.00 &   0.00 & &      $\infty$ &      $\infty$ & $\infty$ &     $\infty$ &     $\infty$ \\
	$f_{17}$  & & 0.00 & 0.00 & 0.00 & 0.00 &   \textbf{0.04} & &      $\infty$ &      $\infty$ & $\infty$ &     $\infty$ & \textbf{4307962} \\
	$f_{18}$  & & 0.00 & 0.00 & 0.00 & 0.00 &   0.00 & &      $\infty$ &      $\infty$ & $\infty$ &     $\infty$ &     $\infty$ \\
	$f_{19}$  & & 0.00 & 0.00 & 0.00 & 0.00 &   0.00 & &      $\infty$ &      $\infty$ & $\infty$ &     $\infty$ &     $\infty$ \\
	\cmidrule(l){1-1} \cmidrule(l){3-7} \cmidrule(l){9-13}
	$f_{20}$  & & 0.00 & 0.00 & 0.00 & 0.00 &   0.00 & &      $\infty$ &      $\infty$ & $\infty$ &     $\infty$ &     $\infty$ \\
	$f_{21}$  & & \textbf{0.57} & 0.02 & 0.00 & 0.11 &   0.22 & &   \textbf{164332} & 11050720 & $\infty$ & 1683405 &  720267 \\
	$f_{22}$  & & \textbf{0.01} & 0.00 & 0.00 & 0.00 &   0.00 & & \textbf{22318143} &      $\infty$ & $\infty$ &     $\infty$ &     $\infty$ \\
	$f_{23}$  & & 0.00 & 0.00 & 0.00 & 0.00 &   0.00 & &      $\infty$ &      $\infty$ & $\infty$ &     $\infty$ &     $\infty$ \\
	$f_{24}$  & & 0.00 & 0.00 & 0.00 & 0.00 &   0.00 & &      $\infty$ &      $\infty$ & $\infty$ &     $\infty$ &     $\infty$ \\
	\cmidrule(l){1-1} \cmidrule(l){3-7} \cmidrule(l){9-13}
	\textbf{Overall} & & \textbf{0.46} & 0.44 & 0.00 & 0.14 & 0.43 & & \textbf{7 bests} & 3 bests & 0 bests & 1 best & 3 bests \\
	\bottomrule
\end{tabular}

    }
    \caption{Success Rates (SRs) and Expected Run Times (ERTs) restricted to the functions with dimension $D=40$ and $0.01$ as target error. Best results per function are denoted in boldface. The last row of the table contains the average SRs and the number of times the algorithm obtained the best ERT. Horizontal lines are used to group the functions with similar characteristics, as described in Section~\ref{sec:expsetup_ben}.}
    \label{tab:target001}
\end{table}

Overall, BHPOP obtains the highest success rate and, most notably, 10 best $\mathit{ERT}$ values are obtained by BHPOP and BH, whereas the rest of the competitor algorithms only obtained four best $\mathit{ERT}$s.
The two Basin Hopping variants seem to be much quicker in the unimodal functions (see in particular $f_1$ and $f_5$).
The best improvement of BHPOP with respect to BH is especially noticeable in functions $f_{21}$ and $f_{22}$.
In particular, BHPOP is the only algorithm with a non-null success rate on $f_{22}$.
Conversely, CMA-ES is the only algorithm able to solve $f_{17}$ in at least one execution, with the considered precision and within the allowed budget of evaluations.

\begin{figure}[H]
    \centering
    \includegraphics[width=1.0\textwidth]{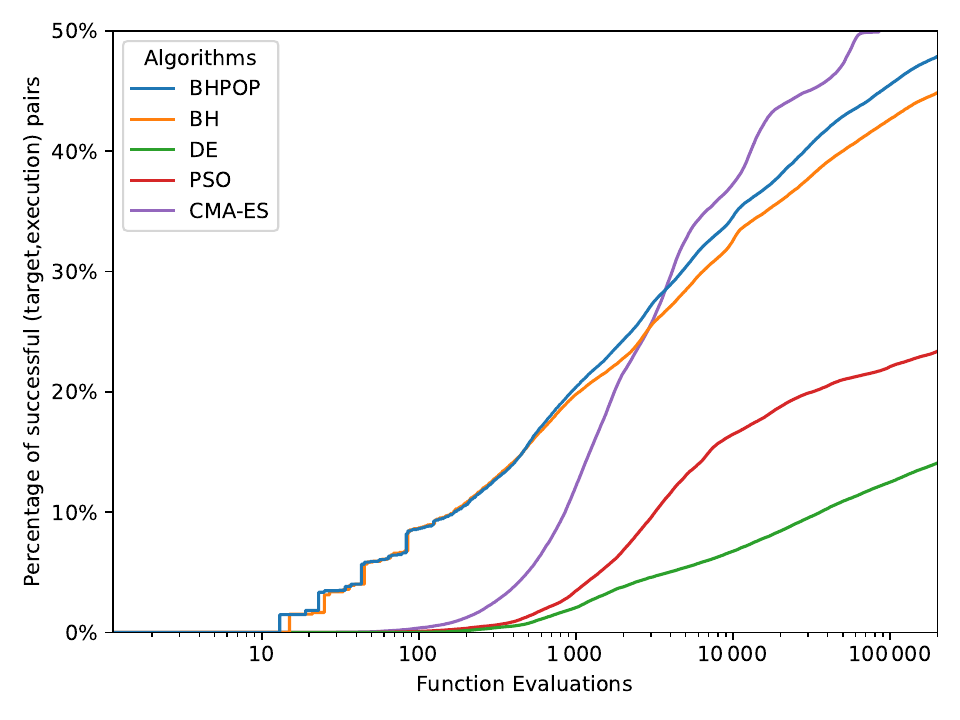}
    \caption{The curves represent the empirical cumulative density function (ECDF) for each algorithm inset as a function of the
    number of function evaluations on the x-axis (log scale). ECDF curves show the increase in percentage of successful (target,execution) pairs
    as the number of function evaluations increases.}
    \label{fig:runtime}
\end{figure}

In order to summarize the analysis with a single picture, Figure~\ref{fig:runtime} plots, for each algorithm, the empirical cumulative density function (ECDF) of the percentage of successful executions, by considering all the five targets $t \in \{ 10^{-8}, 10^{-4}, 0.01, 0.1, 1 \}$, as the number of function evaluations increases.

Figure~\ref{fig:runtime} shows that, on average:
BHPOP starts to improve its success rate with respect to BH after roughly 500 evaluations;
BHPOP and BH are more than competitive with CMA-ES for budgets until about $25\,000$ evaluations; 
while, for larger budgets, Figure~\ref{fig:runtime} confirms the same ``three-bands'' standing derived from Figure~\ref{fig:heatmap200k}, i.e., CMA-ES is more successful than BHPOP and BH, which, in turn, are largely more successful than PSO and DE.

\subsection{Discussion}
\label{sec:expres_disc}

This experimental investigation, performed over a wide range of benchmark functions which cover many different characteristics that may be encountered in real-world problems, shows that the simple search scheme of Basin Hopping may be competitive with more recent and sophisticated metaheuristics, especially when out-of-the-box implementations, available in popular and largely adopted software libraries, are considered.

\begin{figure}[H]
    \centering
    \includegraphics[width=1.0\textwidth]{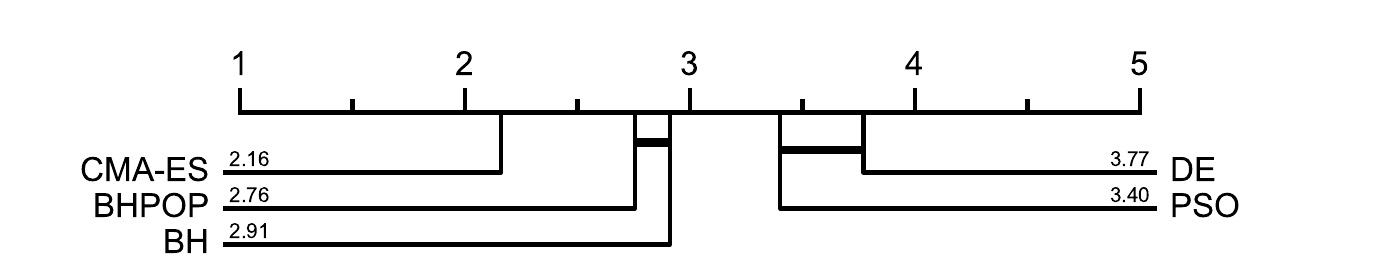}
    \caption{Critical difference diagram showing pairwise statistical difference comparison of the five algorithms over the considered problems, according to the Conover post-hoc test with Benjamini-Hochberg adjustment and $\alpha=0.05$ as significance level. The main horizontal axis represents ranks, the algorithms point to and are labeled with their average ranks, while thick horizontal lines connect cliques of algorithms which are not significantly different to each other.}
    \label{fig:cd_diagram}
\end{figure}

As an extreme synthesis of the analyses provided in Sections~\ref{sec:expres_budget} and~\ref{sec:expres_target}, it is possible to derive a three-bands standing of the five algorithms as follows: CMA-ES /BHPOP, BH / PSO, DE.
In order to have a clearer visual perception of the three-bands standing, we provide in Figure~\ref{fig:cd_diagram} the critical difference diagram\footnote{Critical difference diagrams have been introduced in~\cite{cddiagram1} and then adapted to generic pairwise comparison tests in~\cite{cddiagram2}.} showing the Friedman average ranks of the algorithms together with thick horizontal lines connecting cliques of algorithms that are not significantly different to each other, according to the same adjusted Conover post-hoc procedure used for Figure~\ref{fig:heatmap200k}.

Delving deeper into the details not captured by the high-level summary depicted in Figure~\ref{fig:cd_diagram}, it is interesting to observe where BHPOP outperforms vanilla BH, and when and why it might be preferred to the overall best performing CMA-ES algorithm.

Firstly, the results presented in Section~\ref{sec:expres_target} show that BHPOP improves the success rate of BH across the entire benchmark suite. This suggests that incorporating a population management mechanism into the basic BH search scheme leads to a more robust algorithm.
Secondly, a closer examination of the results provided in Section~\ref{sec:expres_budget} shows that BHPOP significantly improves the performance of vanilla BH when dealing with problems presenting a weakly regular multi-modal landscape. This improvement likely arises from the population-based scheme of BHPOP, which fosters a ``breadth-focused'' search, allowing BHPOP to explore the irregularities of the landscape more extensively compared to the trajectory-based BH algorithm.

Moreover, while CMA-ES emerges as the overall top-performing algorithm based on the experimental results examined so far, it is important to highlight that there are specific scenarios where BHPOP may be a more advantageous choice over CMA-ES.
Indeed, the results presented in Sections~\ref{sec:expres_budget}~and~\ref{sec:expres_target} show that BHPOP outperforms CMA-ES in four scenarios:
  (i)~when the problem dimension is low ($D=5$),
 (ii)~for separable problems, even when they have a higher dimension,
(iii)~when a moderate precision level ($10^{-2}$) is required,
 (iv)~when the available budget is limited to around 1000 evaluations.
Essentially, BHPOP may be preferred to CMA-ES for simpler problems or in situations characterized by tight resource constraints, i.e., those scenarios where the covariance matrix maintained by CMA-ES becomes more of a hindrance than a benefit.




Finally, to investigate the computational time in terms of seconds,
we carried out an extra 
experiment
where we recorded the execution time of BHPOP, CMA-ES and a silly random search algorithm.
The three algorithms were executed $10$ times on the benchmark function $f_{24}$, with a budget of $10\,000$ evaluations, and by considering five increasing dimensions $D \in \{ 20,40,60,80,100 \}$.
All the algorithms are implemented in Python (CMA-ES comes from the previously mentioned Nevergrad library) and this 
experiment
is run on a machine equipped with a CPU Intel Core i7-10510U with a maximum clock rate of $2.30$ GHz, $16$ GB of memory and Windows 10 as operating system.
The average computational time in seconds of BHPOP and CMA-ES are provided in Table~\ref{tab:times}, together with their overheads with respect to the random search execution time (i.e., the time required by BHPOP/CMA-ES divided by the time required by the random search).

\begin{table}
    \centering
    \resizebox{1.0\textwidth}{!}{
        \begin{tabular}{lp{0.2cm}cccccp{0.2cm}ccccc}
\toprule
 & & \multicolumn{5}{l}{\textbf{Time in seconds}} & & \multicolumn{5}{l}{\textbf{Overhead w.r.t. random search}} \\
 \vspace{-0.2cm} \\
\textbf{Algorithm} & & $D=20$ & $D=40$ & $D=60$ & $D=80$ & $D=100$ & & $D=20$ & $D=40$ & $D=60$ & $D=80$ & $D=100$ \\
\cmidrule(l){1-1} \cmidrule(l){3-7} \cmidrule(l){9-13}
BHPOP    & &            0.28 & 0.47 & 0.95 & 1.56 & 2.69 & &                          \phantom{0}1.51 &  \phantom{0}1.95 &  \phantom{0}3.32 &  \phantom{0}3.89 &  \phantom{0}5.58 \\
CMA-ES & &            3.10 & 4.51 & 4.16 & 5.35 & 5.63 & &                         16.87 & 18.76 & 14.54 & 13.41 & 11.69 \\
\bottomrule
\end{tabular}

    }
    \caption{Computational time in seconds and overhead with respect to a pure random search, averaged over 10 executions of $10\,000$ evaluations on $f_{24}$ by considering different dimensions.}
    \label{tab:times}
\end{table}

As expected, Table~\ref{tab:times} clearly shows that BHPOP has a much lower computational complexity than CMA-ES, thus further extending the scenarios where simple Basin Hopping metaheuristics may be considered in practical applications.

\section{Real World Problems}
\label{sec:realworld}

Comparing algorithms on well designed benchmark function sets, as the one used here, is no
doubt a useful step forward towards understanding the origin of problem difficulty and the
behavior of a given metaheuristic. However,
testing the algorithms on some difficult problem coming from a real application field in science or
engineering is also highly recommended as, in the end, all these techniques are developed to solve real problems.
The first problem we consider is the
Lennard-Jones potential used to compute atomic clusters energies which is among the examples
contained in the CEC2011 document describing a number
of difficult optimization problems coming from real applications~\cite{das2010}. Atomic clusters are systems constituted
by tens or at most a few hundreds of atoms held together by forces that are, in a first crude approximation,
 derivable from semi-empirical
two-atom potentials.  Stable cluster structures at zero temperature must be at the minimum of the
potential energy given as the sum of all pair interactions between the cluster atoms. 
Two semi-empirical potentials are often used: the Lennard-Jones potential and the Morse potential.
The interatomic
potential $V_{ij}$ of the Lennard-Jones type between atoms $i$ and $j$ is:
$$
V_{ij} = 4 \epsilon \bigg [\bigg (\frac{\sigma}{r_{ij}}\bigg)^{12} - \bigg(\frac{\sigma}{r_{ij}}\bigg)^{6} \bigg ],
$$
 where $r_{ij}$ is the distance between atoms $i$ and $j$, $\epsilon$ is the depth of the potential well, and
 $\sigma$ is the distance at which the potential is zero. The first term represents a repulsion while the second
 is attractive.
The total potential energy $E$ to minimize for a cluster of $N_a$ atoms is given by: 

$$ E = 4 \epsilon \sum_{i=1}^{N_a-1}  \sum_{j=i+1}^{N_a} V_{ij}. $$ 
In the calculation, reduced units are used setting $\epsilon$ and $\sigma$ to 1.
This is a hard problem that has been tackled with various stochastic techniques giving rise to known putative global minima
for values of $N_a$ up to $N_a\sim200$. The equilibrium structures thus obtained are also very interesting in
Chemical Physics (see, e.g.,~\cite{wales1997,wales1999}) but, given our focus on metaheuristics, will not be discussed here.

While Lennard-Jones potentials come from the study of the interaction of inert gases like Neon, the Morse
potential was conceived for the case of diatomic molecules to study their vibrational behavior. For a cluster of
$N_a$ atoms with two-body interactions only, it has the following form:
$$
E = \epsilon \sum_{i=1}^{N_a-1}  \sum_{j=i+1}^{N_a} \epsilon e^{(\rho(1 - r_{ij}/r_e)}(e^{\rho(1-r_{ij}/r_e)}-2),
$$
\noindent where $r_{ij}$ is the separation between atoms $i$ and $j$, $\epsilon$ is the pair well depth,
$r_e$ is the pair equilibrium distance, and $\rho$ is a dimensionless parameter that determines the range of the
interatomic forces. Here we use reduced parameters $\epsilon=1$, $r_e=1$, and $\rho=6$.
Finding the stable structures, i.e., those having the minimal potential energy for Morse clusters is at least as
hard as in the Lennard-Jones case. All the known best minima for both the Lennard-Jones and Morse potentials, for $N_a>2$ and up to $N_a=150$, are tabulated and can be found at \url{www-wales.ch.cam.ac.uk/CCD.html}.


In keeping with the black-box approach followed in the benchmark using the test functions described in the previous sections,
and to avoid biases against methods that are gradient-free, we do not make direct use of the gradient of the above functions, 
although it can be easily computed analytically. Also, while for best results in large clusters problem knowledge is required, for the sake of comparison our metaheuristics only use the problem objective function. In this way, the
test is more representative of general black-box global optimization.

Table~\ref{tab:rw} summarizes the results obtained on the above problems.
The problem instances are identified by a string of the form $P\_N_a$, where $P \in \{ \mathit{LJ}, \mathit{MO} \}$ indicates Lennard-Jones or Morse, while $N_a \in \{ 20,30,40 \}$ is the number of atoms. Since both the problems require to find three coordinates per atom, the dimension is $D=3N_a$.
The averages are taken over $15$ executions and the number of allowed function evaluations per run is $2 \times 10^4\times D$, the same for all the algorithms tested.
Table~\ref{tab:rw} provides the average, standard deviation and minimal objective value returned by the algorithms.

\begin{table}
    \centering
    \resizebox{1.0\textwidth}{!}{
        \begin{tabular}{crp{0.005cm}rrrrrp{0.005cm}rrrrr}
\toprule
 & & & \multicolumn{5}{l}{\textbf{Average results $\pm$ Std. Deviations}} & & \multicolumn{5}{l}{\textbf{Best results}} \\
\vspace{-0.2cm} \\
\textbf{Problem} & $\bm{D}$ & & \multicolumn{1}{c}{\textbf{BHPOP}} & \multicolumn{1}{c}{\textbf{BH}} & \multicolumn{1}{c}{\textbf{DE}} & \multicolumn{1}{c}{\textbf{PSO}} & \multicolumn{1}{c}{\textbf{CMA-ES}} & & \multicolumn{1}{c}{\textbf{BHPOP}} & \multicolumn{1}{c}{\textbf{BH}} & \multicolumn{1}{c}{\textbf{DE}} & \multicolumn{1}{c}{\textbf{PSO}} & \multicolumn{1}{c}{\textbf{CMA-ES}} \\
\cmidrule{1-2} \cmidrule{4-8} \cmidrule{10-14}
\vspace{-0.2cm} \\
LJ\_20 & 60  & &   $\mbox{-}72.08 \pm 2.83$ &   $\textbf{-73.40} \pm \phantom{0}5.07$ &  $\mbox{-}13.58 \pm 2.35$ &  $\mbox{-}55.10 \pm \phantom{0}9.48$ &   $\mbox{-}68.78 \pm \phantom{0}4.46$ & &  -76.21 &  \textbf{-77.18} & -19.71 &  -69.03 &  -76.21 \\
\vspace{-0.2cm} \\
LJ\_30 & 90  & &  $\textbf{-122.49} \pm 2.36$ &  $\mbox{-}122.05 \pm \phantom{0}6.74$ &  $\mbox{-}15.41 \pm 1.57$ &            $\mbox{-}92.56 \pm 15.44$ &            $\mbox{-}109.13 \pm 28.50$ & & -126.59 & \textbf{-127.75} & -18.64 & -119.77 & -126.16 \\
\vspace{-0.2cm} \\
LJ\_40 & 120 & &  $\mbox{-}174.00 \pm 3.63$ &            $\textbf{-175.73} \pm 11.29$ &  $\mbox{-}19.38 \pm 2.30$ &           $\mbox{-}129.72 \pm 15.23$ &  $\mbox{-}170.75 \pm \phantom{0}3.59$ & & -178.94 & \textbf{-183.15} & -23.62 & -162.76 & -176.78 \\
\vspace{-0.2cm} \\
\cmidrule{1-2} \cmidrule{4-8} \cmidrule{10-14}
\vspace{-0.2cm} \\
MO\_20 & 60  & &   $\mbox{-}33.54 \pm 4.37$ &             $\textbf{-53.33} \pm 10.56$ &  $\mbox{-}11.12 \pm 1.28$ &  $\mbox{-}44.33 \pm \phantom{0}8.30$ &             $\mbox{-}37.57 \pm 10.64$ & &  -43.83 &  \textbf{-70.72} & -13.57 &  -58.34 &  -62.03 \\
\vspace{-0.2cm} \\
MO\_30 & 90  & &   $\mbox{-}57.03 \pm 5.08$ &             $\textbf{-78.91} \pm 13.14$ &  $\mbox{-}13.09 \pm 1.22$ &            $\mbox{-}70.53 \pm 12.08$ &             $\mbox{-}58.24 \pm 15.71$ & &  -65.02 & \textbf{-100.39} & -15.51 &  -93.71 &  -74.14 \\
\vspace{-0.2cm} \\
MO\_40 & 120 & &   $\mbox{-}84.56 \pm 7.96$ &            $\textbf{-119.19} \pm 20.03$ &  $\mbox{-}16.07 \pm 1.44$ &            $\mbox{-}95.88 \pm 13.89$ &             $\mbox{-}88.76 \pm 13.03$ & & -102.30 & \textbf{-151.74} & -18.60 & -122.60 & -114.06 \\
\vspace{-0.2cm} \\
\cmidrule{1-2} \cmidrule{4-8} \cmidrule{10-14}
\vspace{-0.2cm} \\
\multicolumn{2}{c}{\textbf{Average Ranks}} & &
\multicolumn{1}{c}{2.83} &
\multicolumn{1}{c}{\textbf{1.17}} &
\multicolumn{1}{c}{5.00} &
\multicolumn{1}{c}{3.00} &
\multicolumn{1}{c}{3.00} & &
\multicolumn{1}{r}{3.17} &
\multicolumn{1}{r}{\textbf{1.00}} &
\multicolumn{1}{r}{5.00} &
\multicolumn{1}{r}{3.17} &
\multicolumn{1}{r}{2.67} \\
\bottomrule
\end{tabular}

    }
    \caption{Average results, standard deviations and best results obtained by the five algorithms on the six real-world problems. In the first two columns LJ and MO stand for Lennard-Jones and Morse respectively, and $D$ is the problem dimension which is equal to $3 \times$ number of atoms in the cluster.
    For each problem, the result of the best algorithm is provided in bold, while the last line contains the average ranks over the six problems here considered.}
    \label{tab:rw}
\end{table}

\begin{figure}
    \centering
    \begin{subfigure}[b]{0.49\textwidth}
        \includegraphics[width=\textwidth]{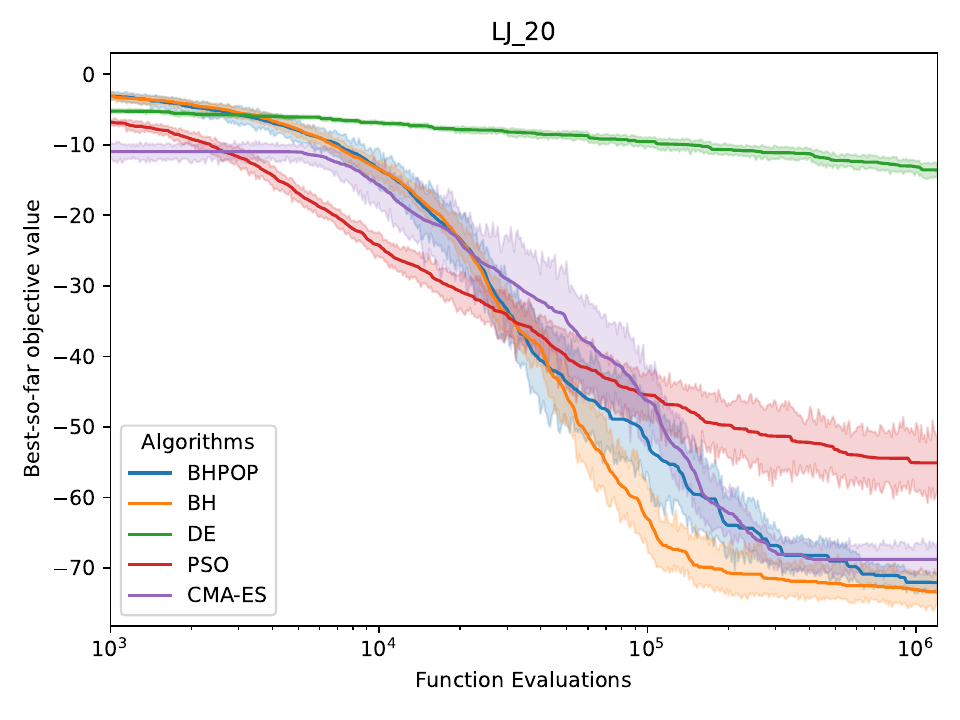}
    \end{subfigure}
    \hspace{-0.1cm} 
    \begin{subfigure}[b]{0.49\textwidth}
        \includegraphics[width=\textwidth]{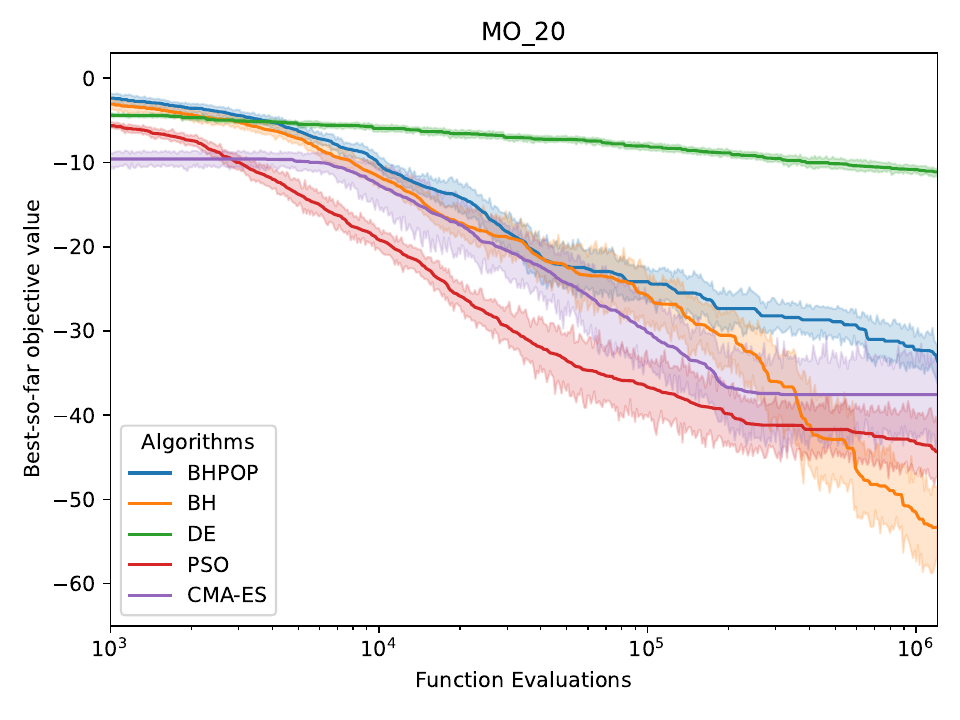}
    \end{subfigure}
    
    \begin{subfigure}[b]{0.49\textwidth}
        \includegraphics[width=\textwidth]{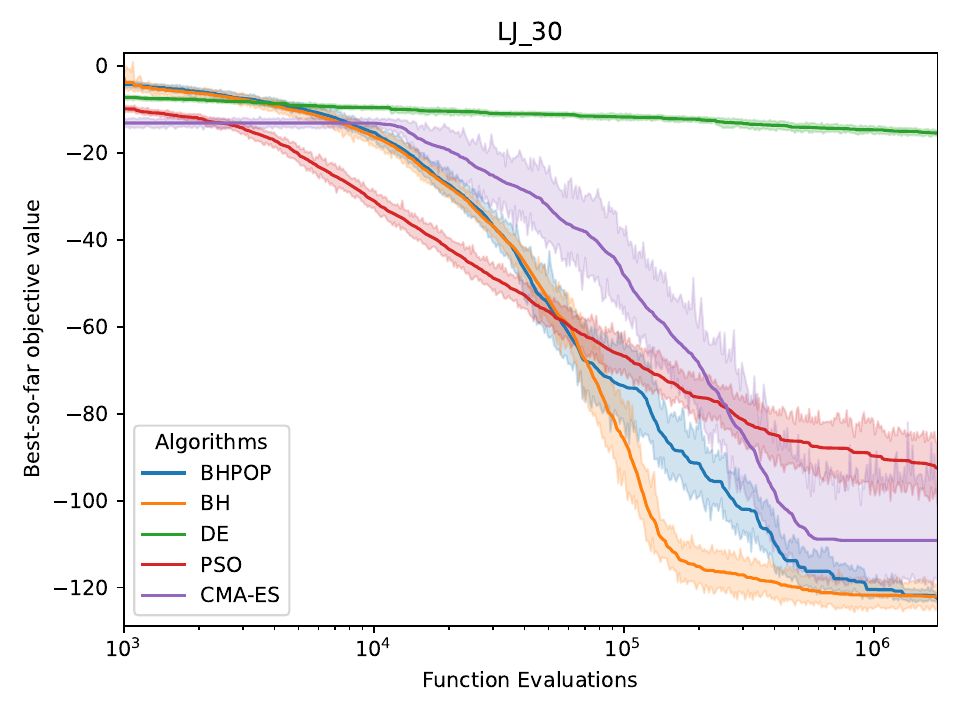}
    \end{subfigure}
    \hspace{-0.1cm} 
    \begin{subfigure}[b]{0.49\textwidth}
        \includegraphics[width=\textwidth]{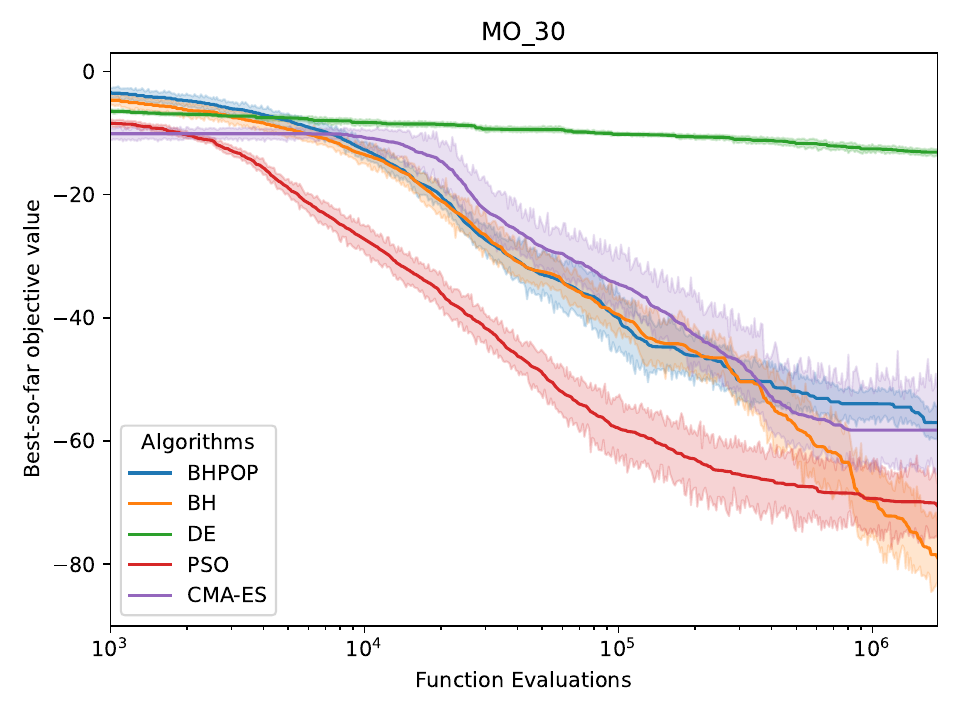}
    \end{subfigure}
    
    \begin{subfigure}[b]{0.49\textwidth}
        \includegraphics[width=\textwidth]{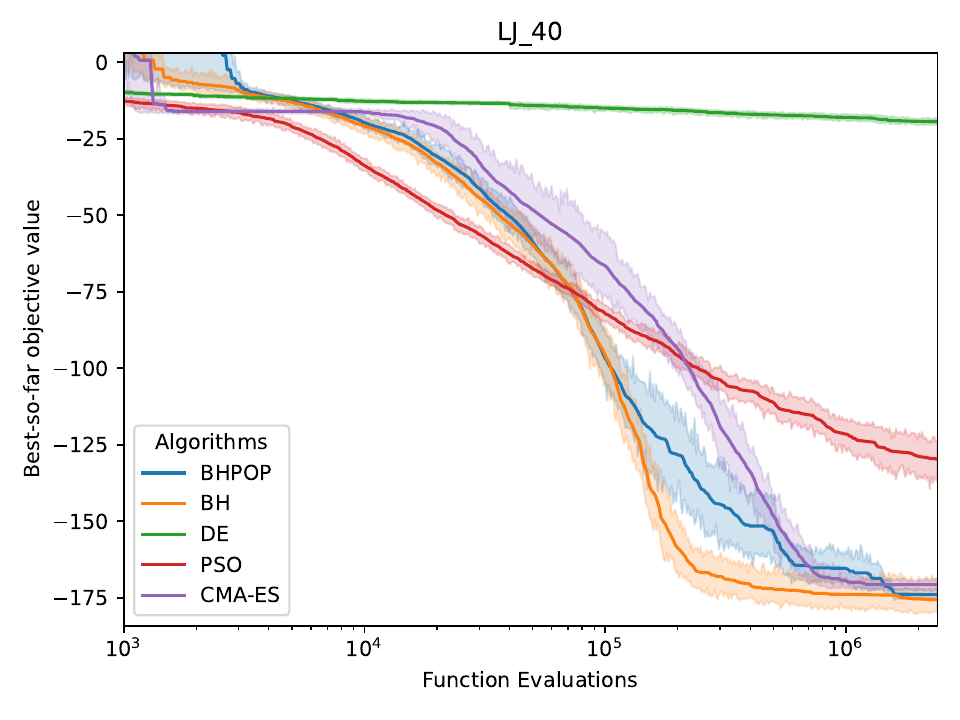}
    \end{subfigure}
    \hspace{-0.1cm} 
    \begin{subfigure}[b]{0.49\textwidth}
        \includegraphics[width=\textwidth]{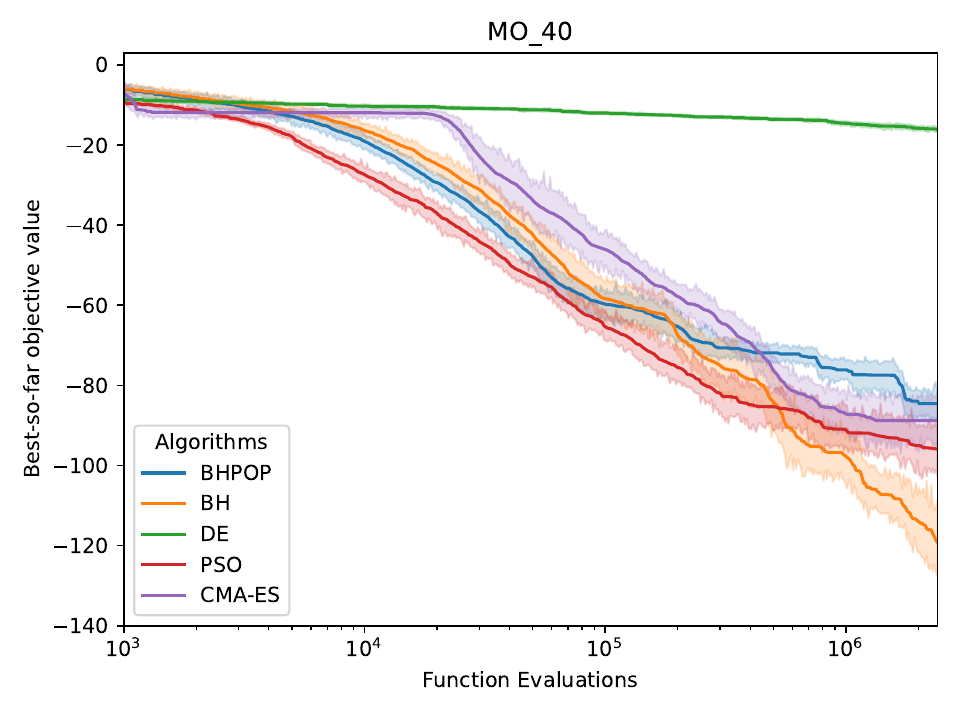}
    \end{subfigure}
    
    \caption{Convergence curves of the best-so-far objective value of the five algorithms for the Lennard-Jones and Morse cluster problems with 20, 30, and 40 atoms. The curves are averaged over $15$ runs for each problem. The shaded area represents the 95\% confidence interval. The x-axis is in log-scale and, for the sake of presentation, it is shown starting from 1000 evaluations.}
    \label{fig:convgraphs}
\end{figure}

The results indicate that BH obtains the best results overall on the two
problems, both on average as well as absolute best.  BHPOP is a close second on the LJ problems, but it
is less accurate on the Morse problems, where it is beaten by CMA-ES and PSO.
DE obtains the worse results on all problems and problem sizes.
The ranking confirms BH, BHPOP and CMA-ES as being the best methods on both synthetic and real-world functions,
at least as far as the function test set and the two problems studied here are concerned. It is of course possible that more sophisticated and finely tuned versions
of PSO and DE would be more competitive but our philosophy throughout the paper was to use entry-level metaheuristics that
are easy to parameterize and run.

Figure~\ref{fig:convgraphs} shows another aspect of the computational experiment, namely the convergence curves for the two problems and the three problem sizes considered. It is apparent that BH not only outperforms the other algorithms in terms of accuracy but it is also, in general, the fastest one. Depending on the problem and its size, BHPOP is very good on LJ but it is slower on the Morse problems, on which CMA-ES and PSO show slightly better performances. 
However, it is interesting to note that the convergence curves of BH and BHPOP almost overlap up to about $100\,000$ evaluations while, after this point, BH seems to boost its convergence towards a better region of the search space. Then, in the LJ problems BHPOP fills the gap after about $1\,000\,000$ evaluations, while this does not happen in the more difficult MO problems (at least, with the allocated budget). Apparently, the population-based approach of BHPOP slows down the convergence of the basin hopping search strategy for this type of problems.

\section{Conclusions and Future Work}
\label{sec:concl}

Our aim in this study was to compare a straightforward optimization algorithm called Basin Hopping and
a newly introduced population variant of it (called BHPOP),  with 
the popular metaheuristics DE, PSO, and CMA-ES.
These metaheuristics are used daily with
satisfactory results by many researchers and we were thus interested in finding out whether 
BH can be competitive with them. To be of any value, a comparison such as this must be carried out on
a significant benchmark and the results analyzed for statistical significance. To fulfill these requirements 
we performed  numerical experiments
using the \textit{IOH profiler} environment with the BBOB test function set.
The experiments were carried out in two different but complementary ways: we measured the performance under a given
fixed budget of function evaluations and by considering a fixed target value. The former gives information about the
algorithm's accuracy, i.e., how close it comes to the known global optimum, while the second focuses on the number
of evaluations required to reach a given target objective function value.

We can summarize the results as follows. On the accuracy criterion at a fixed budget, three algorithms, i.e., CMA-ES, BHPOP
and BH, are significantly superior to PSO and DE, with some differences on different function types. Among the
best three, CMA-ES is the clear winner, closely followed by BHPOP and BH. In the other class, PSO is better overall than
DE.  
The results of the fixed target analysis confirm that the same three algorithms stand out with PSO, and especially DE being less efficient.
Beyond this high-level summary, we have also observed that: (i) the population-based approach of BHPOP improves the performances of the vanilla BH scheme, especially in weakly regular multi-modal landscapes, and (ii) there are specific scenarios where BHPOP can be preferred to CMA-ES, i.e., when the problem at hand is separable or low-dimensional and under tight time constraints or when a moderate accuracy suffices.
It must be said once again that the results are strictly valid only for the BBOB test function set and for the
algorithms with the parameterization used here. It might therefore well be, for example, that a better version of DE, which had the worst results here, would outperform some or all of the other metaheuristics.
However, we think that the results do show a clear general trend and that they are relevant in the scenario where
out-of-the-box algorithms' implementations provided in a well known software library are adopted.

To strengthen our analysis, we also used two hard real-world problems: the minimization of the potential
energy of atomic clusters, where atoms are held together by atom-atom potentials of the Lennard-Jones and Morse 
types. The results on three medium-size instances of each problem showed that BH gives the best results, followed by BHPOP, CMA-ES, PSO, and DE.

The general conclusion, as far as BH and BHPOP are concerned, is that they are almost as good than
CMA-ES on the synthetic benchmark functions and better than it on the two hard cluster energy minimization
problems. Thus, BH can be considered a good candidate, especially if one wants to obtain quick and
reliable results on an unknown problem. On the other hand, it is probably more difficult to
improve on the basic version of BH with respect to more sophisticated approaches contained in evolved versions of
PSO and DE.

Future work includes a deeper study of the perturbation technique in BH and alternative
methods for selection in BHPOP, as well as the role of the population size with respect to the problem
dimension. 
In fact, in light of the convergence analysis performed on the two real-world problems considered in this work, we believe that a dynamically changing population size can allow the search to: (i)~reduce the population size and speed up the convergence once a good region of the search space has been located and, conversely, (ii)~increase the population size in order to escape from stagnation situations when they occur.
Finally, we would also like to extend the 
experiments
to other benchmark test sets, other recent metaheuristics not examined here, and the inclusion of more real-world problems.



\end{document}